\documentclass[runningheads]{llncs}

\usepackage[mobile]{eccv}

\usepackage{eccvabbrv}

\usepackage{graphicx}
\usepackage{booktabs}

\usepackage[accsupp]{axessibility}  
\usepackage{multirow}
\usepackage{booktabs}
\usepackage{tikz}

\usepackage{comment}
\usepackage{amsmath,amssymb}

\usepackage{orcidlink}
\usepackage{color}
\usepackage{booktabs} 
\definecolor{citeblue}{rgb}{0.21,0.49,0.74}
\usepackage[accsupp]{axessibility}  
\usepackage[dvipsnames]{xcolor}
\usepackage{colortbl}
\usepackage{nicematrix}
\usepackage{hyperref}
\usepackage{wrapfig}
\usepackage{picinpar}
\usepackage{cutwin}

\usepackage{tikz}
\usepackage{lipsum}
\usepackage{paralist}
\usepackage{color}

\newcommand*\circlegray[1]{\tikz[baseline=(char.base)]{
\node[shape=circle,draw=gray,fill=gray,inner sep=0.6pt] (char) {\textcolor{white}{\footnotesize \textbf{#1}}};}}

\usepackage{hyperref}

\usepackage{orcidlink}

\makeatletter
\renewcommand*{\@fnsymbol}[1]{\ensuremath{\ifcase#1\or *\or \dagger\or \ddagger\or
    \mathsection\or \mathparagraph\or \|\or **\or \dagger\dagger
    \or \ddagger\ddagger \else\@ctrerr\fi}}
\makeatother

\begin{document}

\title{Referring Atomic Video Action Recognition} 

\titlerunning{RAVAR}

\author{Kunyu Peng\inst{1,}\thanks{Equal contribution} \orcidlink{0000-0002-5419-9292}
\and Jia Fu\inst{3,4,*}\orcidlink{0009-0004-3798-8603}
\and Kailun Yang\inst{2,}\thanks{Correspondence: kailun.yang@hnu.edu.cn}\orcidlink{0000-0002-1090-667X} 
\and Di Wen\inst{1}\orcidlink{https://orcid.org/0009-0000-1693-7912}
\and Yufan Chen\inst{1}\orcidlink{0009-0008-3670-4567}
\and\\Ruiping Liu\inst{1}\orcidlink{0000-0001-5245-2277}
\and Junwei Zheng\inst{1}\orcidlink{0009-0005-4390-3044}
\and Jiaming Zhang\inst{1}\orcidlink{0000-0003-3471-328X}
\and M. Saquib Sarfraz\inst{1,6}
\and\\Rainer Stiefelhagen\inst{1}\orcidlink{0000-0001-8046-4945}
\and Alina Roitberg\inst{5}\orcidlink{0000-0003-4724-9164}
}

\authorrunning{K.~Peng, J. Fu et al.}

\institute{$^1$Karlsruhe Institute of Technology, $^2$Hunan University,\\$^3$RISE Research Institutes of Sweden, $^4$KTH Royal Institute of Technology,\\$^5$University of Stuttgart, $^6$Mercedes-Benz Tech Innovation}

\maketitle

\begin{abstract}
We introduce a new task called \textbf{R}eferring \textbf{A}tomic \textbf{V}ideo \textbf{A}ction \textbf{R}ecognition (RAVAR), aimed at identifying atomic actions of a particular person based on a textual description and the video data of this person. This task differs from traditional action recognition and localization, where predictions are delivered for all present individuals. In contrast, we focus on recognizing the correct atomic action of a specific individual, guided by text. To explore this task, we present the RefAVA dataset, containing $36,630$ instances with manually annotated textual descriptions of the individuals. 
To establish a strong initial benchmark, we implement and validate baselines from various domains, \textit{e.g.}, atomic action localization, video question answering, and text-video retrieval. Since these existing methods underperform on RAVAR, we introduce \texttt{RefAtomNet} -- a novel cross-stream attention-driven method specialized for the unique challenges of RAVAR: the need to interpret a textual referring expression for the targeted individual, utilize this reference to guide the spatial localization and harvest the prediction of the atomic actions for the referring person. The key ingredients are: (1) a multi-stream architecture that connects video, text, and a new location-semantic stream, and (2) cross-stream agent attention fusion and agent token fusion which amplify the most relevant information across these streams and consistently surpasses standard attention-based fusion on RAVAR. Extensive experiments demonstrate the effectiveness of \texttt{RefAtomNet} and its building blocks for recognizing the action of the described individual. The dataset and code will be made publicly available at \href{https://github.com/KPeng9510/RAVAR}{RAVAR}.
\end{abstract}

\section{Introduction}
Referring scene understanding~\cite{liu2017referring,yuan2021instancerefer,liu2019clevr,qiu2020language} aims to solve the underlying computer vision task for the particular scene element specified via a natural language \textit{referring expression}.
Context-driven queries are vital in information retrieval, assistive systems, and multimedia analysis, but incorporating them into computer vision models is challenging due to a complex entanglement of linguistic descriptions, localization, and visual recognition itself.
Several benchmarks for referring scene understanding have been successfully established for multi-object tracking~\cite{wu2023referring}, semantic segmentation~\cite{khoreva2019video,shi2023unsupervised,li2018referring,shi2018key,seo2020urvos}, medical imaging~\cite{seibold2022reference}, and object detection~\cite{dang2023instructdet,pramanick2022doro}.%
While these efforts focus on object-centric referring expressions, many applications require a human-centric understanding of human actions, \eg, in rehabilitation assistance~\cite{saha2018fine,laput2019sensing} and human-robot interaction~\cite{lea2016learning,ji2019context}.
Our work addresses a new problem --  reference-driven recognition of atomic actions performed by the human described via a textual reference, which has been overlooked in the past.

The field of human action recognition has made exciting advances, from new approaches stemming from the rise of visual transformers~\cite{ryali2023hiera,wang2023videomae,li2022mvitv2,wang2023masked,wang2022internvideo,peng2022transdarc,gritsenko2023end} to new large-scale datasets~\cite{carreira2017quo,goyal2017something,kuehne2011hmdb,soomro2012ucf101,shao2020finegym}.
Datasets utilized for the prediction of atomic actions often feature recordings with multiple individuals~\cite{gu2018ava}.
Nevertheless, the vast majority of previously published works~\cite{kim2024atrous,ryali2023hiera,wang2023videomae,gritsenko2023end,rajasegaran2023benefits} either rely on manually window crop of a particular person of interest or automatically generated region of interests to predict the atomic actions for all the individuals in one video, which results in large pre- or post-processing effort to focus on one specific person.%
\begin{figure*}[t!]
\centering
\includegraphics[width=1\linewidth]{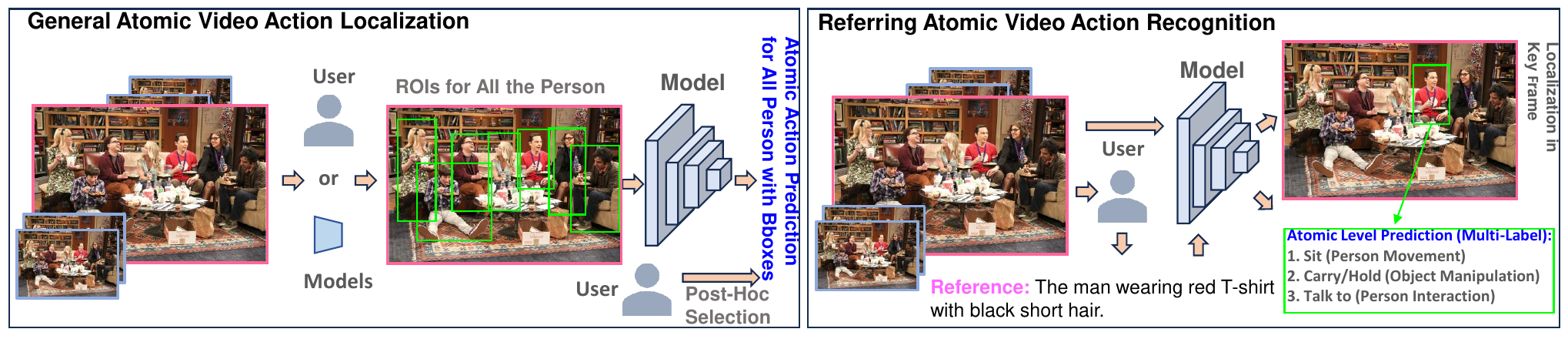}
\caption{A comparison of AAL Task (left) and RAVAR task (right).}
\label{fig:task_comparison}
\end{figure*}
This workflow is also very inconvenient in certain applications, such as assistive systems for users with visual impairments~\cite{zheng2023materobot,liu2023open,ou2022indoor}, who must understand the state of each present in the scene for effective interactions.
Leveraging succinct textual descriptions, which may include broad positional indicators (\textit{e.g.}, left, center, and right), appearance attributes (\textit{e.g.}, hair color and clothing), or gender, to steer the model towards delivering atomic action recognition outcomes for the targeted individual, presents a promising solution to the outlined challenges. 
Employing these textual cues as a reference, the model can facilitate an end-to-end retrieval of the specified instance across the entire video, subsequently providing precise atomic action recognition results.

To close this gap, we formalize the new task of \textbf{R}eferring \textbf{A}tomic \textbf{V}ideo \textbf{A}ction \textbf{R}ecognition (RAVAR). The differences between RAVAR and atomic action localization are shown in Fig.~\ref{fig:task_comparison}. 
The conventional atomic action recognition in multi-person scenarios is often formalized in an action localization manner, and requires the region of interest to localize the human available a priori, after which atomic action recognition is carried out for each individual. Post-processing is required for individuals of interest. 
In contrast, RAVAR assumes a textual reference and a video as inputs, delivering the referring individual's atomic actions along with the location. 
This comparison shows the efficiency of RAVAR, leveraging textual information for less labor-intensive subject identification to assist atomic action analysis for the individual of interest. We further introduce the new RefAVA dataset for RAVAR.
RefAVA is established using the $17,946$ video clips of the public AVA dataset~\cite{gu2018ava} initially designed for atomic action localization, which we extend with $36,630$ textual descriptions of the corresponding individuals used as the referring expression.
The dataset covers diverse settings, \eg, in/out door and day/night time scenarios, and numerous multi-person scenes, constituting an ideal testbed for our task.
To establish an initial benchmark, we assess $15$ well-established approaches from multiple related research domains.
Our experiments reveal that none of the benchmarked methods delivers sufficient performance. 
For example, previous methods within the domains of VQA and VTR typically yield predictions of a coarse analysis regarding actions, whereas strategies in AAL face challenges in spatially localizing the correct referring individual.
We attribute this phenomenon to the massive irrelevant information provided by the visual data, distracting the model from the referring location. 
How to suppress the referring irrelevant information becomes a critical challenge to pursue better RAVAR performances.
To address this, we introduce the new \texttt{RefAtomNet} approach specialized in simultaneously solving the tasks of understanding textual \texttt{ref}erring expressions and using them as guides for spatial localization and classification of \texttt{atom}ic actions. 
We begin by computing textual, visual, and newly proposed \textit{location-semantic tokens} using a large foundation model and well-established object detector.  
The latter stream is designed to harvest semantic cues of different scene entities: it leverages a pre-trained object detector to localize and identify objects, after which the location-semantic tokens are computed as a fusion of visual bounding box coordinates and text embeddings of the object category. 
Another novel aspect of our work is the introduction of \textit{cross-stream agent attention and agent token fusions}, which builds on the concept of agent attention~\cite{han2023agent} and re-defines it for cross-stream compatibility: agent tokens are reformulated into a 1D sequence format through the use of fully connected layers, bypassing the initial need for 2D pooling, removing the depthwise convolution branch, and forgoing 2D positional encodings. This mechanism enhances the model's ability to filter and emphasize relevant information from multiple data sources effectively.
\texttt{RefAtomNet} achieves the highest performance for RAVAR, while a detailed ablation study demonstrates the effectiveness of each component.

Our contributions can be summarized as follows:
\begin{itemize}
   \item We introduce the new \textbf{R}eferring \textbf{A}tomic \textbf{V}ideo \textbf{A}ction \textbf{R}ecognition (RAVAR) task and the RefAVA benchmark with $36,630$ atomic action instances manually annotated with suitable language expressions.  
    \item To establish a competitive benchmark, we examine $11$ well-established models from different related fields, namely atomic action localization, video question answering, and video-text retrieval. 
    \item We propose \texttt{RefAtomNet}, a new RAVAR approach that uses a novel agent-based semantic-location aware attentional fusion to integrate multi-modal tokens, suppressing the irrelevant visual cues. \texttt{RefAtomNet} delivers improvements of $3.85\%$ and $3.17\%$ of mAP, and $4.33\%$ and $4.03\%$ of AUROC, on val and test sets, respectively, compared with the leading baseline, BLIPv2~\cite{li2023BLIP}.

\end{itemize}
\section{Related Work}
\noindent\textbf{Referring Scene Understanding.}
Referring scene understanding aims to locate parts of interest within images or videos guided by natural language, exhibiting the utility in many computer vision tasks including autonomous driving~\cite{wu2023referring} and video editing~\cite{chai2023stablevideo}. 
The development of this field cannot be advanced without the contribution of high-quality open-source datasets and benchmarks~\cite{bu2022scene,yu2016modeling, vasudevan2018object, khoreva2019video, seo2020urvos, wu2023referring, dang2023instructdet,lin2024echotrack}.
For example, the CLEVR-Ref+ benchmark is proposed by Liu~\etal~\cite{liu2019clevr} to achieve visual reasoning with referring expressions. Li~\etal~\cite{li2018referring} proposed referring image segmentation by using a recurrent refinement network. Wu~\etal~\cite{wu2023referring} proposed referring multi-object tracking benchmark. 
However, there are no referring understanding works focusing on atomic video action analysis, whereupon we first conduct RAVAR benchmarking with our RefAVA dataset. We further clarify that Referring Video Object Segmentation (RVOS) task~\cite{liu2021cross,su2023sequence,gavrilyuk2018actor,mcintosh2020visual} differs us from multiple perspectives, \textit{e.g.}, RVOS includes action name in input textual reference. 

\noindent\textbf{Video Text Retrieval.}
Video Text Retrieval (VTR) aims at matching relevant video content with text. 
Amidst the rise of Vision-Language foundation models like CLIP~\cite{radford2021learning} and BLIP~\cite{li2022BLIP}, efforts~\cite{wang2023actionclip,luo2022clip4clip, ma2022x, li2023BLIP, zhang2023multi,wu2023cap4video,chen2023tagging,madasu2023improving,lin2023towards,shi2023learning} are being made to apply powerful pre-trained models for promoting their competencies in VTR.
XCLIP~\cite{ma2022x} introduced multi-grained contrastive learning for end-to-end VTR, by counting all the video-sentence, video-word, sentence-frame, and frame-word contrasts.
BLIPv2~\cite{li2023BLIP} pre-trained a Querying Transformer (QFormer) to bootstrap vision-and-language representation learning and vision-to-language generative learning. The key difference between VTR and RAVAR lies in their inference tasks: VTR queries text or video to localize elements, while RAVAR integrates textual and visual data for fine-grained subject retrieval and atomic action prediction in videos.

\noindent\textbf{Video Question Answering.}
Video Question Answering (VQA) focuses on generating answers to questions posed in natural language for a given video. 
Depending on the emphasis of the question, factoid VQA~\cite{yang2021just, castro2022wild,gao2023mist,liu2023cross,li2023lavender,chen2023video,zhang2023video,bagad2023test,le2020hierarchical,jiang2020divide,xiao2021next,garcia2020knowit,lei2021less,guo2021re} straightforward queries visual facts. 
Lei~\textit{et al.}~\cite{lei2022revealing} proposed that a single-frame trained transformer-based model, with large-scale pre-training and a frame ensemble at the inference stage, can perform better than existing multi-frame trained models in factoid VQA tasks. 
On the other hand, inference VQA~\cite{li2022representation, gandhi2022measuring} delves into logical reasoning. 
Li~\textit{et al.}~\cite{2023videochat} released a video-centric instruction dataset and leveraged a neural interface to integrate video foundation models and Large Language Models (LLM), showcasing capability in temporal reasoning, causal inference, and event localization.
In addition, some multimodal VQA frameworks~\cite{li2022learning, garcia2020knowit, yang2022avqa} explore information-invoking scenarios that incorporate visual, audio, subtitle, and external knowledge.
However, compared with the RAVAR task, most of the existing VQA approaches do not particularly focus on atomic action recognition and will only tend to give a coarse textual description, which limits the applications requiring precise prediction, \eg, human-robot assistance~\cite{lea2016learning,ji2019context}.

\noindent\textbf{Atomic Video Action Recognition and Localization.}
Atomic video-based action recognition~\cite{gritsenko2023end,chung2021haa500} and localization~\cite{gu2018ava} involves the identification and analysis of the most fundamental, indivisible actions or movements performed by humans for single and multiple-person scenarios. Compared with the general video action recognition task, atomic video-based action localization is much more fine-grained and is always formulated in a multi-label manner with bounding box predictions. Most of the existing convolutional neural networks (CNN)~\cite{carreira2017quo,wang2023videomae,feichtenhofer2020x3d,feichtenhofer2019slowfast} and the transformer~\cite{ryali2023hiera,wang2023videomae,li2022mvitv2,wang2023masked,wang2022internvideo,peng2022transdarc,gritsenko2023end} networks for human action recognition are commonly used in the atomic video-based action localization by changing the classification head into multi-label manner, adding additional bounding box prediction head, and integrating region of interest features from human detector. 
Ryali~\textit{et al.}~\cite{ryali2023hiera} proposed a hierarchical vision transformer with high efficiency and precision within the realm of the existing methods using video as input. 
Wang~\textit{et al.}~\cite{wang2023videomae} proposed scaling video-masked autoencoders with dual masking.
Current methods for predicting individual actions in multi-person scenarios often require manual video cropping for atomic action recognition or generate predictions for all detected individuals, necessitating further human selection and reducing practicality~\cite{pramono2021spatial,wang2023stal,rajasegaran2023benefits}. Most existing AAL methods are not specifically designed for the RAVAR task.
We thereby introduce \texttt{RefAtomNet}, a novel method that utilizes location semantics atop predicted location and scene semantic information derived from the scenario together with the textual reference and visual cues, then further uses cross-stream agent attention and agent token fusions to suppress redundant information.

\section{RAVAR: Established Benchmark}
\label{sec:3-0}
\begin{table}[t]
\caption{
An overview of the referring scene understanding datasets and our RefAVA dataset;
Our task is Referring Atomic Video Action Recognition (RAVAR), whereas existing benchmarks focus on Referring Object Detection (ROD), Referring Video-based Object Segmentation (RVOS), and Referring Multi-Object Tracking (RMOT).}
\label{tab:datasets}

\scalebox{0.54}{\begin{tabular}{l|lllllllll}
\toprule
\midrule
\textbf{Dataset}       & RefCOCO~\cite{yu2016modeling}                 & RefCOCO+~\cite{yu2016modeling}                & RefCOCOg~\cite{yu2016modeling}                & Talk2Car~\cite{deruyttere2019talk2car} & VID-Sentence~\cite{chen2019weakly} & Refer-DAVUS17~\cite{khoreva2019video} & Refer-YV~\cite{zeng2022motr} & Refer-KITTI~\cite{wu2023referring} & \textbf{RefAVA}    \\
\midrule
\textbf{Task}& ROD & ROD & ROD & ROD & ROD          & RVOS  & RMOT     & RMOT        & \textbf{RAVAR}     \\
\textbf{N$_{Frames}$}   & 26,711  & 19,992                  & 26,711 & 9,217    & 59,238       & 4,219         & 93,869  & 6,650 & \textbf{1,615,140} \\
\textbf{N$_{Instance}$} & 26,711  & 19,992  & 26,711  & 10,519   & 7,654  & 3,978  & 7,451    & -     & \textbf{36,630}\\
\midrule
\bottomrule
\end{tabular}}

\end{table}

\subsection{Introduction of the RefAVA Dataset}
\label{sec:3-1}

\noindent\textbf{Textual Annotations.} To acquire precise textual annotations for the individuals of interest, $7$ annotators manually provided the textual annotations according to the key frame bounding boxes presented in the AVA dataset~\cite{gu2018ava}. 
Cross-checking among all the annotators for the annotated individuals is conducted to deliver high textual annotation quality.

\begin{figure}[t!]
    \centering
    \begin{minipage}{0.53\textwidth}
        \centering
        \includegraphics[width=0.95\linewidth]{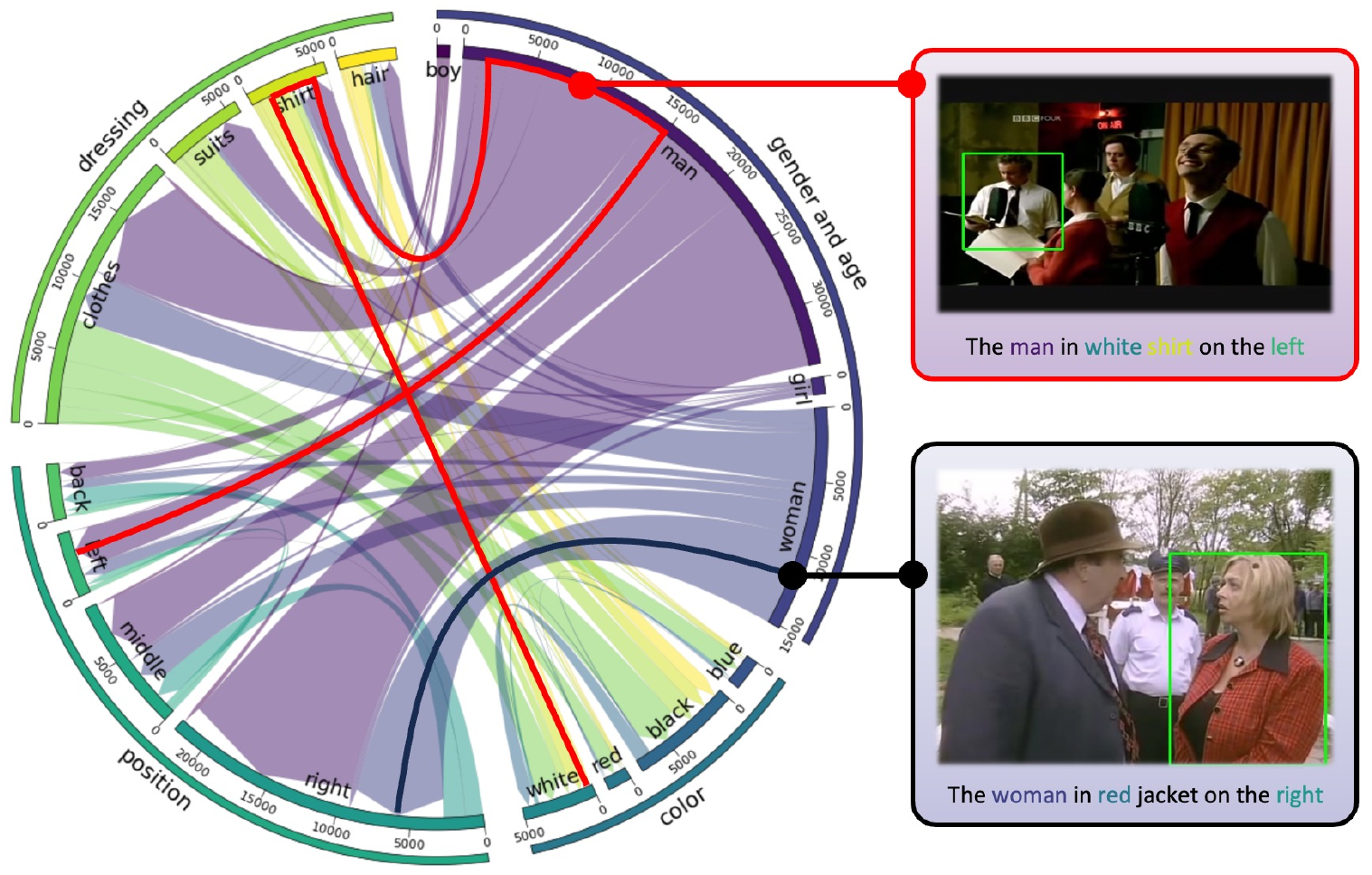}
        \subcaption{Chord visualization of several keywords.}
        \label{fig:chord}
    \end{minipage}%
    \begin{minipage}{0.47\textwidth}
        \centering
        \includegraphics[width=1\linewidth]{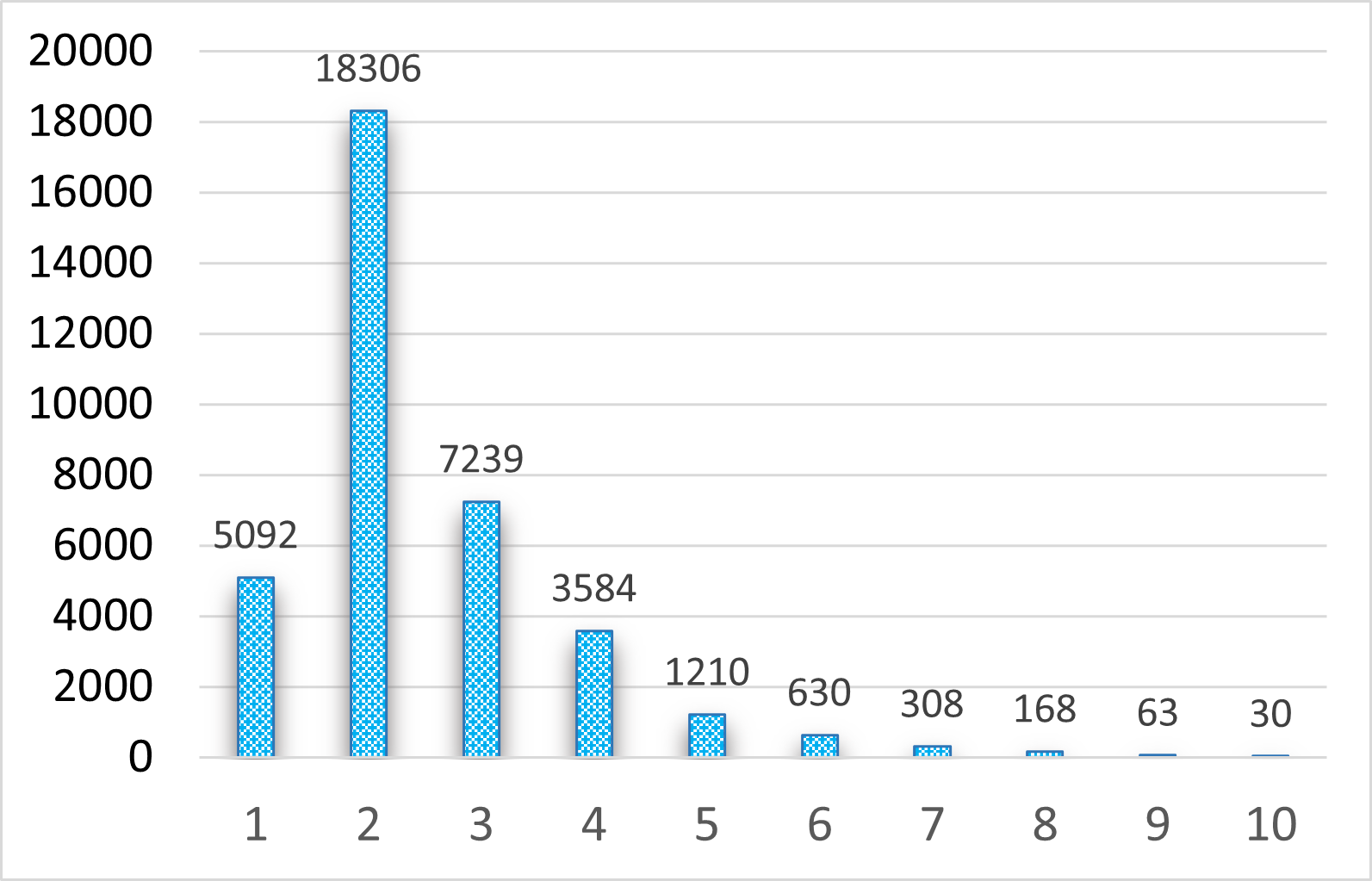}
        \subcaption{%
Annotation statistics: \#The number of annotated persons contained in the scenario (x-axis); \#The number of instances (y-axis).}
        \label{fig:statistcs_per_video}
    \end{minipage}

\caption{An chord visualization of several keywords from the textual references, shown on the left, and the instance amount in video clips containing different numbers of annotated persons in our RefAVA dataset, shown on the right.}

\label{fig:statistics}
\end{figure}

\noindent\textbf{Dataset.} We selected $17,946$ video clips from $127$ movies of the AVA dataset, preserving the most complex scenarios, and annotated each person on each center frame of every $90$s video clip. In total, RefAVA has $36,630$ labeled instances, and we split them into $22,658$ train instances, $10,916$ validation instances, and $3,056$ test instances, where the samples from different sets are from different movie scenarios. 
The textual annotations cover information on approximate age, gender, appearance, relative position in the center frame, \textit{etc.}, and without action description.
The chord visualization in Fig.~\ref{fig:chord} reveals the quantitative causal relations between the most frequent notional words in different categories from the textual references. 
Further, Fig.~\ref{fig:statistcs_per_video} shows the statistics of instance amount towards the annotated person number inside the scenario where each instance belongs. 
We deliver the comparison of the statistics between representative existing referring scene understanding datasets and our RefAVA dataset in Tab.~\ref{tab:datasets}. The atomic actions involve $80$ categories covering Object Manipulation (OM), Person Interactions (PI), and Person Movement (PM). The videos cover diverse scenarios. The test set is sourced from $26$ movies different from those in the train set ($67$ different movies) and val set ($34$ different movies) to achieve the evaluation of generalizability.

\subsection{The Baselines for the RAVAR Benchmark}
\label{sec:3-2}
We have adapted methodologies from similar fields as our baselines for RAVAR.

\noindent\textbf{AAL Baselines.} We first reformulate the existing methods from the general atomic video action localization field by integrating the textual reference embeddings into the visual branch. Some action recognition approaches, \eg, I3D~\cite{carreira2017quo} and X3D~\cite{feichtenhofer2020x3d}, are adapted to AAL following~\cite{feichtenhofer2019slowfast}. BERT~\cite{Devlin2019BERTPO} is used for textual embedding extraction. 
The selected baselines from this domain can be grouped into CNN-based approaches~\cite{carreira2017quo,feichtenhofer2020x3d} and transformer-based approaches~\cite{li2022mvitv2,ryali2023hiera,wang2023videomae}. All these approaches leverage the pre-trained weight on the Kinetics400~\cite{carreira2017quo}. 

\noindent\textbf{VQA Baselines.} The second group of approaches comes from the video question answering domain, which contains the GPT-based model, \ie, AskAnything~\cite{2023videochat}, and conventional transformer-based model~\cite{lei2022revealing}. 
We input the reference sentence in a questioning manner and add a classification head and a bounding box regression head after the acquisition of the logits. 
We finetuned the two models for the RAVAR task based on their pre-trained weight on ActivityNetQA~\cite{yu2019activitynet}. 

\noindent\textbf{VTR Baselines.} The last group of the foundation model baselines comes from the video-text retrieval task, where XCLIP~\cite{ma2022x}, CLIP4CLIP~\cite{luo2022clip4clip}, BLIPv2~\cite{li2023BLIP}, and MeVTR~\cite{zhang2023multi} are selected. These foundation models are pre-trained on a combination of numerous datasets, incorporating Conceptual Captions~\cite{sharma2018conceptual}, SBU Captions~\cite{ordonez2011im2text}, and COCO Captions~\cite{chen2015microsoft}, \textit{etc}. 
We reformulate these approaches in the same way as the approaches in VQA.

\noindent\textbf{SF Baselines.} Some Single Frame (SF) baselines are adopted, where SAM~\cite{kirillov2023segment}, DETR~\cite{carion2020end}, and REFCLIP~\cite{jin2023refclip} are utilized together with the CLIP~\cite{radford2021learning} feature extraction backbone.

\noindent\textbf{VOS Baseline.} We adopt one Video Object Segmentation (VOS) baseline using the encoder of the approach from Su~\textit{et al.}~\cite{su2023sequence} and our prediction head.

\section{RefAtomNet: Proposed Method}
\subsection{Overview}
We propose a new model, \texttt{RefAtomNet}, with an overview provided in Fig.~\ref{fig:main}.
It leverages three token streams: visual, textual reference, and location-semantic streams. Visual tokens are extracted from the video using ViT~\cite{dosovitskiy2020vit}, while textual reference tokens are obtained from the text using BERT~\cite{Devlin2019BERTPO}. Both streams are enhanced by QFormer~\cite{li2023BLIP}. Additionally, in the location-semantic stream, location-semantic tokens are extracted from the center frame by fusing predicted object coordinates and semantic embeddings of the object categories estimated by a frozen object detector and BERT~\cite{Devlin2019BERTPO}. 
To suppress irrelevant information for each stream, we propose the agent-based location-semantic aware attentional fusion to achieve better amplification of relevant information during the cross-stream exchange.
The detailed structure is discussed in the following subsections.

\subsection{RefAtomNet}
\noindent\textbf{Background of QFormer.} We rely on BLIPv2 to extract the visual features, where QFormer is the most essential component. Multiple learnable queries are initialized in QFormer as trainable parameters and interact with input data via Transformer attention mechanisms. Each query selectively attends to input parts, capturing task-specific features and updating based on attention outputs. During training, these queries are optimized for specialized information extraction. Queries then directly contribute to generating outputs.

\noindent\textbf{Extraction of Visual and Textual Reference Tokens.} 
Similar to BLIPv2~\cite{li2023BLIP}, we use a pre-trained multimodal model for token extraction from both the video and textual reference streams. This model leverages the QFormer~\cite{li2023BLIP} architecture to effectively extract and integrate multimodal embeddings from the visual and textual reference data. Specifically, the visual stream employs a ViT~\cite{dosovitskiy2020vit} backbone encoder to process visual inputs, while the textual branch utilizes BERT~\cite{Devlin2019BERTPO} to encode textual reference cues. This combined approach facilitates a holistic understanding of both visual content and textual descriptions.
The model extracts visual tokens ($\mathbf{t}^{VT}$) and textual reference tokens ($\mathbf{t}^{RT}$) through Eq.~\ref{eq:1}:
\begin{equation}
\label{eq:1}
    \mathbf{t}^{VT}, \mathbf{t}^{RT} = \mathcal{V}_{VL}(\mathcal{V}_{VT}(\mathbf{x}^{VT}),~ \mathcal{V}_{RT}(\mathbf{x}^{RT})), 
\end{equation}
where $\mathcal{V}_{VL}$ represents the visual-textual integration model (\textit{i.e.}, QFormer~\cite{li2023BLIP}). $\mathcal{V}_{VT}$, and $\mathcal{V}_{RT}$ denote the backbones for extracting visual tokens (\textit{i.e.}, ViT~\cite{dosovitskiy2020vit}) and textual reference tokens (\textit{i.e.}, BERT~\cite{Devlin2019BERTPO}), respectively. $\mathbf{x}^{VT}$ and $\mathbf{x}^{RT}$ represent the input video and textual reference caption, which are then fed into liner projection layers, \ie, $\mathbf{P}_{VT}$ and $\mathbf{P}_{RT}$, respectively.

\noindent\textbf{Extraction of Location-Semantic Aware Tokens.}
To integrate more location and semantic information into the token representations, we leverage a well-established object detector DETR~\cite{carion2020end} to deliver $N_o$ detection results based on the input of the center frame of a video clip, noted as the keyframe. Note that, the detections provided by the object detector contain either the human or other objects with a high confidence score according to Eq.~\ref{eq_2}. 
\begin{equation}
\label{eq_2}
    \mathbf{r}_{boxes}, \mathbf{r}_{cats} = \mathcal{V}_{dets}(\mathbf{x}_k),
\end{equation}
where $\mathbf{x}_k$, $\mathcal{V}_{dets}$ indicates the keyframe and the detection network. $\mathbf{r}_{boxes}\in \mathbb{R}^{N_o \times 4}$ are detected 2D corner coordinates (top left and bottom right corners) of the bounding boxes. $\mathbf{r}_{cats}$ is the predicted category for each bounding box represented in textual format. These category labels of the detected objects are passed to a text encoder (BERT~\cite{Devlin2019BERTPO}) to extract semantic embeddings.
We concatenate these two types of tokens together along the channel dimension and use a single linear projection layer to obtain the aggregated location-semantic aware tokens ($\mathbf{t}^{LS}$) as Eq.~\ref{eq:3}.
\begin{equation}
\label{eq:3}
    \mathbf{t}^{LS} = \mathbf{P}_{LS}(Concat\left[\mathcal{V}_{RT}(\mathbf{r}_{cats}), \mathbf{r}_{boxes}\right]),
\end{equation}
where $\mathbf{P}_{LS}$ indicates a fully connected layer, $\mathcal{V}_{RT}$ indicates the language feature extraction backbone, \ie, BERT~\cite{Devlin2019BERTPO}, and $Concat$ indicates the concatenation.

\noindent\textbf{Agent-Based Location Semantic Aware Attentional Fusion.} 
\begin{figure*}[t!]
\centering
\includegraphics[width=1\linewidth]{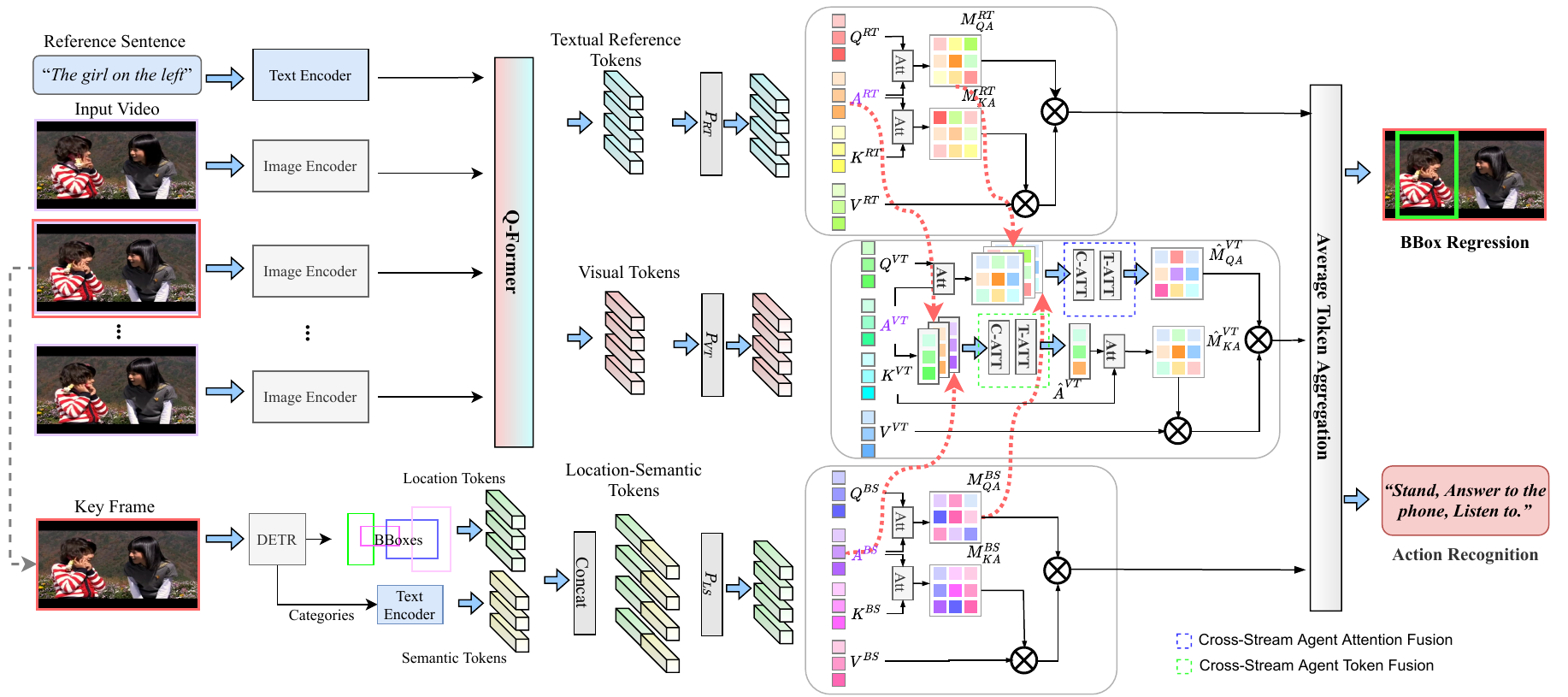}
\caption{An overview of the \texttt{RefAtomNet} architecture.}
\label{fig:main}
\end{figure*}

\noindent\underline{Agent Tokens and Sequential Processing.}
We use \textit{agent tokens}, inspired by agent attention~\cite{han2023agent}. Derived through fully connected layers from input tokens, similar to Query, Key, and Value extraction in transformers, agent tokens aggregate essential information via agent-key and agent-query pairs.
To enable agent attention for 1D sequential tokens from our streams, we redefine the 2D agent spatial tokens proposed in~\cite{han2023agent} into a 1D sequential format for multi-stream fusion. Specifically, we replace 2D pooling with linear projection and eliminate the depthwise convolutional branch and biased position encoding to better suit agent acquisition in our model.
These 1D sequential agent tokens serve dual functions: aggregating crucial information from agent-key and agent-query pairs and efficiently redistributing this information to the original value tokens, therefore, enhancing attention focus and reducing irrelevant cues.

\noindent\underline{Agent-Based Attention Mechanism.}
The agent-based attention mechanism for visual, textual reference, and location-semantic streams is computed as:
\begin{equation}
    \label{eq:4}
    \mathbf{Q}^{\phi}, \mathbf{K}^{\phi}, \mathbf{V}^{\phi}, \mathbf{A}^{\phi} = \mathbf{W}_{Q}^{\phi}(\mathbf{t}^{\phi}), \mathbf{W}_{K}^{\phi}(\mathbf{t}^{\phi}), \mathbf{W}_{V}^{\phi}(\mathbf{t}^{\phi}), \mathbf{W}_{A}^{\phi}(\mathbf{t}^{\phi}),
\end{equation}
where for better readability, we use $\phi$ to represent $\left[RT, VT, LS\right]$, where RT indicates the reference tokens, VT indicates the visual tokens, and LS indicates the location-semantic tokens.
$\mathbf{Q}^{\phi}$, $\mathbf{K}^{\phi}$, $\mathbf{V}^{\phi}$, and $\mathbf{A}^{\phi}$ are the query, key, value, and agent tokens.
$t^{\phi}$ depicts the input. $\mathbf{W}_{Q}^{\phi}$, $\mathbf{W}_{K}^{\phi}$, $\mathbf{W}_{V}^{\phi}$, and $\mathbf{W}_{A}^{\phi}$ are constructed by linear projection layers. We then project the agent tokens using fully connected layers $\mathbf{P}^{\phi}_A$ through $\mathbf{A}_{*}^{\phi}= \mathbf{P}^{\phi}_{A}(\mathbf{A}^{\phi})$ as aforementioned.
To mitigate redundancy cues of the textual reference and location-semantic streams, we leverage agent attention for both of these two streams. 
The agent query attention mask $\mathbf{M}^{\pi}_{QA}$ and the agent key attention mask $\mathbf{M}^{\pi}_{KA}$ are obtained by matrix multiplication ($MatMul$) and SoftMax operations along the channel dimension (indicated by $\sigma_c$) as shown in Eq.~\ref{eq:5}, where $\pi \in \left[RT,LS\right]$ and $\alpha$ indicates a fixed scale factor.
The agent attention masks and tokens are computed and refined to ensure that only pertinent information influences the attention mechanism, as shown in Eq.~\ref{eq:5} and Eq.~\ref{eq:6}:
\begin{align}
    \label{eq:5}
    \mathbf{M}^{\pi}_{QA}, \mathbf{M}^{\pi}_{KA} &= \sigma_c(MatMul[\alpha*\mathbf{A}_{*}^{\pi}, \mathbf{Q}^{\pi}]), \sigma_c(MatMul[\alpha*\mathbf{A}_{*}^{\pi}, \mathbf{K}^{\pi}]), \\
    \label{eq:6}
    \mathbf{t}_{*}^{\pi} &= FFN(MatMul[\mathbf{M}^{\pi}_{KA}, MatMul[\mathbf{M}^{\pi}_{QA}, \mathbf{V}^{\pi}]]),
\end{align}
where $FFN$ indicates the Feed Forward Network. These steps ensure precise model attention, enhancing the contextual relevance of the resulting tokens.

\noindent\underline{Cross-Stream Agent Attention and Agent Token Fusions.}
Finally, cross-stream agent attention and agent token fusion are proposed, which are explicitly tailored for the visual stream by applying the agent query attention maps and the agent tokens from the other two streams. This process involves recalculating the attention maps, thereby ensuring that the final token representation is highly relevant and contextually enriched.
We compute the agent query attention map for the visual token stream, where $\gamma$ = $VT$, as shown in Eq.~\ref{eq:7},
\begin{equation}
\label{eq:7}
    \mathbf{M}^{\gamma}_{QA} = \sigma_c(MatMul\left[\alpha*\mathbf{A}_{*}^{\gamma}, \mathbf{Q}^{\gamma}\right]).
\end{equation}
Then, we calculate the cross-stream irrelevance-suppressed attention for the agent query attention of the visual stream through Eq.~\ref{eq:8} to achieve cross-stream agent attention fusion to suppress irrelevant information in the agent query attentions of the visual stream, the operations used in the second and third terms are abbreviated as C-ATT and T-ATT in Fig.~\ref{fig:main}, 

\begin{equation}
\label{eq:8}
    \hat{\mathbf{M}}^{\gamma}_{QA} =  AVG\left[\mathbf{M}^{\gamma}_{QA} , \sigma_c(\sum_{\pi}\mathbf{M}^{\pi}_{QA}) *  \mathbf{M}^{\gamma}_{QA}  ,\sigma_{t}(\sum_{\pi}\mathbf{M}^{\pi}_{QA}) *  \mathbf{M}^{\gamma}_{QA}\right], 
\end{equation}
where $\sigma_t$ denotes the SoftMax operation along the token dimension. $AVG$ indicates the mean operation.
We follow the same procedure to calculate the cross-stream irrelevance-suppressed agent tokens before the calculation of the agent key attention, as in Eq.~\ref{eq:9} to achieve cross-stream agent token fusion. 

\begin{equation}
\label{eq:9}
    \hat{\mathbf{A}}^{\gamma}_{*} =  AVG \left[\mathbf{A}^{\gamma}_{*} + \sigma_c(\sum_{\pi}\mathbf{A}^{\pi}_{*}) *  \mathbf{A}^{\gamma}_{*}  + \sigma_{t}(\sum_{\pi}\mathbf{A}^{\pi}_{*}) *  \mathbf{A}^{\gamma}_{*}\right], 
\end{equation}
then, we calculate the agent key attention as shown in Eq.~\ref{eq:10},
\begin{equation}
\label{eq:10}
    \hat{\mathbf{M}}^{\gamma}_{KA} = \sigma_c(MatMul\left[\alpha*\hat{\mathbf{A}}_{*}^{\gamma}, \mathbf{K}^{\gamma}\right]).
\end{equation}
The final aggregated visual tokens can be obtained through the following equation as demonstrated in Eq.~\ref{eq:11},
\begin{equation}
\label{eq:11}
    \mathbf{t}_{*}^{\gamma} = FFN(MatMul\left[\hat{\mathbf{M}}^{\gamma}_{KA},MatMul\left[\hat{\mathbf{M}}^{\gamma}_{QA},\mathbf{V}^{\gamma}\right]\right]).
\end{equation}
Finally, we aggregate all the tokens from three branches by using the mean operation, where $N_s$ indicates the number of the stream as shown in Eq.~\ref{eq:12},
\begin{equation}
\label{eq:12}
    \mathbf{t}_{agg} = \sum_{\phi}\left[\mathbf{t}_{*}^{\phi}\right]/N_{s}.
\end{equation}
We construct MLP-based classification and regression heads atop aggregated tokens for center frame bounding box prediction and atomic action recognition.

\subsection{Loss Functions}
We use Binary Cross Entropy (BCE) loss and Mean Squared Error (MSE) loss for the multi-label supervision and the bounding box regression supervision in the same way as all the baselines. The equation of the BCE loss is as Eq.~\ref{eq:13},
\begin{equation}
\label{eq:13}
    L_{BCE}(\mathbf{y}, \hat{\mathbf{y}}) = -\frac{1}{N_c} \sum_{i=1}^{N_c} [\mathbf{y}_i \log(\hat{\mathbf{y}}_i) + (1 - \mathbf{y}_i) \log(1 - \hat{\mathbf{y}}_i)],
\end{equation}
where $\mathbf{y}$, $\hat{\mathbf{y}}$ and $N_c$ indicate the one-hot ground truth, the prediction, and the category number. The bounding boxes regression loss is expressed as Eq.~\ref{eq:14},
\begin{equation}
\label{eq:14}
    L_{MSE}(\mathbf{b}, \hat{\mathbf{b}}) = \sum_{j=1}^{4} (\mathbf{b}_{j} - \hat{\mathbf{b}}_{j})^2,
\end{equation}
where $\mathbf{b}$ indicates the coordinates of the left top and right bottom corners, and $\hat{\mathbf{b}}$ represents the predicted boxes. $j$ indicates the coordinate index.
\section{Experiments}
\subsection{Implementation Details}
We conduct experiments on four NVIDIA A100 GPUs. We use BertAdam~\cite{kingma2014adam} optimizer with learning rate $lr = 1e^{-4}$, batch size $128$, learning rate decay $0.9$, and warmup ratio $0.1$. The agent number and head number in our \texttt{RefAtomNet} are $4$ and $1$.
Our model has $214M$ trainable parameters. The weight for $L_{MSE}$ and $\alpha$ are $5$ and $0.125$. We trained our model for $40$ epochs on our dataset.
 The text encoder is frozen during training. 
We use multi-label mean Average Precision (mAP), Area Under the Receiver Operating Characteristic curve (AUROC), and the mean Intersection over Union (mIOU) as metrics, of which mIOU has less priority since we prioritize the atomic video action recognition task.  

\subsection{Experimental Results}
\noindent\textbf{Main Results.} 
Results of the main experiments on the proposed RefAVA benchmark are summarized in Tab.~\ref{tab:ravar_benchmark}. 
We compare our RefAtomNet model with a
multitude of published approaches adopted to suit our task (see Section ~\ref{sec:3-2}) stemming from the fields of (1) Atomic Action Localization (AAL); (2) Video Question Answering (VQA); (3) Video-Text Retrieval (VTR); (4) Single Frame (SF) baselines; and (5) Video Object Segmentation (VOS).
Baseline methods from the AAL group consistently underperform in spatially localizing the action according to the textual reference (measured as mIOU). 
Even though the ability of the human spatial localization is not the major expected output, it can still generally illustrate if the model can grasp the correct referring person or not. 
Among the AAL methods, X3D~\cite{feichtenhofer2020x3d} achieves the highest AUROC score of $59.09\%$ and $64.51\%$ on the val and test sets, respectively. 
The VQA and VTR baselines work better than the AAL baselines since they can reasonably capture the referring person while delivering acceptable atomic human action recognition performances, which benefits from the text-aware pretraining of these two tasks. AskAnything~\cite{2023videochat} from the VQA group achieves $20.09\%$ and $22.35\%$ of mIOU, $51.42\%$ and $52.25\%$ of mAP, and $66.12\%$ and $69.35\%$ of AUROC on the val and test sets. BLIPv2~\cite{li2022BLIP} from the VTR group delivers $32.99\%$ and $32.75\%$ of mIOU, $52.13\%$ and $53.19\%$ of mAP, and $66.56\%$ and $69.92\%$ of AUROC, on the val and test sets, respectively. 
Our \texttt{RefAtomNet} achieves state-of-the-art performances,  outperforming the best baseline BLIPv2~\cite{li2023BLIP} by $5.23\%$, $3.85\%$, $3.17\%$ and $3.67\%$, $4.33\%$, $4.03\%$ of mIOU, mAP, and AUROC, on val and test sets.
\begin{table}[t!]
\centering
\caption{Experimental results on our RAVAR benchmark.}

\label{tab:ravar_benchmark}
\resizebox{\textwidth}{!}{
\setlength\tabcolsep{8.0pt}
\renewcommand{\arraystretch}{0.7}
\begin{tabular}{r|c|ccc|ccc}
\toprule
\midrule
\multicolumn{2}{c|}{\multirow{2}{*}{\textbf{Method}}}  & \textbf{mIOU} & \textbf{mAP} & \textbf{AUROC} & \textbf{mIOU} & \textbf{mAP} & \textbf{AUROC}  \\ 
\cmidrule{3-8}
\multicolumn{2}{c|}{}  & \multicolumn{3}{c|}{Val}           & \multicolumn{3}{c}{Test} \\ 
\cmidrule{1-8}
\multirow{5}{*}{AAL} &I3D~\cite{carreira2017quo} & 0.00 &44.04&57.77&0.00& 44.64 & 62.71\\
&X3D~\cite{feichtenhofer2020x3d}  &0.26 & 44.45& 59.09& 0.27&46.34&64.51\\
&MViTv2-B~\cite{li2022mvitv2}  & 0.76 &42.32 &56.43 &0.66 &42.60 &59.30 \\
&VideoMAE2-B~\cite{wang2023videomae} & 0.16& 42.02& 55.12& 0.29& 41.87& 59.10\\
&Hiera-B~\cite{ryali2023hiera}  & 0.49& 42.74& 56.72& 0.62& 41.14 &58.70 \\
\cmidrule{1-8}
\multirow{2}{*}{VQA} &Singularity~\cite{lei2022revealing}&26.34&42.43 & 59.52 & 29.78&42.18& 56.12 \\
&AskAnything13B~\cite{2023videochat}  & 20.09 &  51.42  & 66.12  & 22.35  &  52.25 & 69.35\\
\cmidrule{1-8}
\multirow{4}{*}{VTR} &MeVTR~\cite{zhang2023multi}   & 30.78 & 38.42 & 51.01 & 29.79&  36.27&  52.45\\ 
&CLIP4CLIP~\cite{luo2022clip4clip} & 34.75&39.48 &52.57 &32.33&37.17&55.05\\
&XClIP~\cite{ma2022x}   &35.54& 42.46 & 56.30 & 31.79 & 40.82& 58.71\\
&BLIPv2~\cite{li2023BLIP}  & 32.99 & 52.13& 66.56& 32.75& 53.19& 69.92 \\
\midrule
\multirow{3}{*}{SF} 
&CLIP~\cite{radford2021learning}+SAM~\cite{kirillov2023segment} & 32.95 & 49.74 & 64.55 & 29.80 & 51.34 & 69.11 \\
&CLIP~\cite{radford2021learning}+DETR~\cite{carion2020end} & 33.96 & 47.57 & 61.83 & 33.84 & 50.92 & 67.29 \\
&CLIP~\cite{radford2021learning}+REFCLIP~\cite{jin2023refclip} & 34.43 & 47.79 & 62.50 & 32.28 & 49.90 & 67.44 \\
\midrule

\multirow{1}{*}{VOS} 
&Su \textit{et al.}~\cite{su2023sequence} & 23.71 & 52.17 & 66.67 & 26.02 & 53.20 & 70.19 \\
\cmidrule{1-8}
\multicolumn{2}{c|}{RefAtomNet (Ours)}  & \textbf{38.22} & \textbf{55.98} & \textbf{69.73} & \textbf{36.42} & \textbf{57.52} & \textbf{73.95}\\
\midrule
\bottomrule
\end{tabular}
}
\end{table}
\begin{table}[t!]
\footnotesize

\caption{Experiments for module ablation and comparison with multi-modal fusion.}

                \begin{minipage}{0.5\textwidth}
                \raggedright

\subcaption{Module ablation of the RefAtomNet.}

\label{tab:module_ablation}
\scalebox{0.63}{\begin{tabular}{l|ccc|ccc}

\toprule
\midrule
\multirow{2}{*}{\textbf{Method}} & \textbf{mIOU}   & \textbf{mAP}& \textbf{AUROC}&\textbf{mIOU}& \textbf{mAP} & \textbf{AUROC} \\
\cmidrule{2-7}
& \multicolumn{3}{c}{Val} & \multicolumn{3}{|c}{Test}\\
\midrule
w/o ALSAF  &27.30 &50.70  &65.31  &29.09  &51.26 &68.20 \\
w/o CAAF  &36.21 &55.43&69.66 &35.38 & 55.90& 72.60\\
w/o CATF & 35.01 & 53.83& 67.71&34.55 & 56.96& 73.40\\
w/o LSAS  &31.90 & 55.21 &69.47 & 31.25 & 55.73 & 72.38 \\
\midrule
Ours & \textbf{38.22} & \textbf{55.98} & \textbf{69.73} & \textbf{36.42} & \textbf{57.52} & \textbf{73.95}\\
\midrule
\bottomrule
\end{tabular}}

\end{minipage}
\begin{minipage}[c]{0.5\textwidth}

\subcaption{Comparison with other fusion approaches.}

\label{tab:mm}
\scalebox{0.56}{\begin{tabular}{l|ccc|ccc}
\toprule
\midrule
\multirow{2}{*}{\textbf{Fusion}} & \textbf{mIOU}   & \textbf{mAP}& \textbf{AUROC}&\textbf{mIOU}& \textbf{mAP} & \textbf{AUROC} \\
\cmidrule{2-7}
& \multicolumn{3}{c}{Val} & \multicolumn{3}{|c}{Test}\\
\midrule
Addition   &27.30 &50.70  &65.31  &29.09  &51.26 &68.20 \\
Concatenation  &18.64 & 52.23 &66.45 & 20.65& 53.44& 70.70\\
Multiplication  & 23.90 & 51.55 & 65.66& 25.05 & 53.33 & 70.48\\
AttentionBottleneck~\cite{nagrani2021attention}  &33.47  & 50.97 &65.07 &33.02 &54.02  &71.08 \\
McOmet~\cite{zong2023mcomet}  &23.88  & 51.58 &65.65 &25.02  & 53.21 &70.42 \\
\midrule
Ours  & \textbf{38.22} & \textbf{55.98} & \textbf{69.73} & \textbf{36.42} & \textbf{57.52} & \textbf{73.95}\\
\midrule
\bottomrule
\end{tabular}}
\end{minipage}%

\end{table}

\noindent\textbf{Ablations of the Individual Modules.} In Tab.~\ref{tab:module_ablation}, we show the ablation for the components of \texttt{RefAtomNet} by removing each of the proposed designs, where LSAS indicates the location-semantic stream, CAAF indicates the cross-stream agent attention fusion, CATF indicates the cross-stream agent token fusion, and \textit{w/o} ALSAF indicates by simply using addition to fuse these three streams and without the agent-based location-semantic aware attentional fusion mechanism. Compared with the ablation \textit{w/o} ALSAF, \texttt{RefAtomNet} achieves promising improvements of $10.92\%$, $5.28\%$, $4.42\%$ and $7.33\%$, $6.26\%$, $5.75\%$ in terms of mIOU, mAP, and AUROC for the val and test sets, respectively. 
It indicates that using simple aggregation of the three streams makes the model distract from the important cues for the referring person, illustrated by the large decay of mIOU metric. 
Compared with the ablation \textit{w/o} LSAS, we find that the location-semantic tokens benefit more for the localization of the correct person in the center frame. 
The advantages of the LSAS on atomic video action recognition metrics are highlighted in the test set which has higher scenario diversity compared with the val set. 
Using CAAF and CATF, each of them brings promising benefits by suppressing the irrelevant information in the visual tokens.

\noindent\textbf{Comparison with Other Fusion Mechanisms.}
We compare our proposed agent-based location-semantic aware attentional fusion mechanism with late fusion addition, multiplication, concatenation, AttentionBottleNeck~\cite{nagrani2021attention}, and McOmet~\cite{zong2023mcomet} in Tab.~\ref{tab:mm}. 
Our approach outperforms all by suppressing those irrelevant visual tokens on the agent tokens and attention masks.

\noindent\textbf{Generalizability to Test-time Reference Rephrasing.}
\begin{table}[t]
\footnotesize
\caption{Generalizability to different referring styles and encoder architectures.}

\begin{minipage}{0.5\textwidth}
\raggedright

\subcaption{Generalizability to test time rephrasing}

\label{tab:rephrase}
\scalebox{0.61}{\begin{tabular}{l|ccc|ccc}
\toprule
\midrule
\multirow{2}{*}{\textbf{Fusion}} & \textbf{mIOU}   & \textbf{mAP}& \textbf{AUROC}&\textbf{mIOU}& \textbf{mAP} & \textbf{AUROC} \\
\cmidrule{2-7}
& \multicolumn{3}{c}{Val} & \multicolumn{3}{|c}{Test}\\
\midrule
Singularity~\cite{lei2022revealing}  & 18.45 & 41.27 & 58.47 & 20.54 & 41.39 & 55.32\\
XCLIP~\cite{ma2022x}  & 31.95 & 41.35  & 54.35  & 29.84  & 40.74  & 58.45 \\
AskAnything~\cite{2023videochat}  & 19.74 & 51.11 & 65.83& 21.60 & 51.96 & 69.04\\
BLIPv2~\cite{li2023BLIP}  & 31.00 & 51.45 & 65.88 & 31.34 & 52.35 & 68.87 \\
\midrule
Ours  & \textbf{34.65} & \textbf{55.75} &\textbf{69.52} & \textbf{33.14}  & \textbf{57.23}  &\textbf{73.76} \\
\midrule
\bottomrule
\end{tabular}}

\end{minipage}
\begin{minipage}[c]{0.5\textwidth}
\subcaption{Generalizability to different visual-textual encoder architectures.}
\label{tab:gen_tvr}
\scalebox{0.57}{\begin{tabular}{l|ccc|ccc}
\toprule
\midrule
\multirow{2}{*}{\textbf{Fusion}} & \textbf{mIOU}   & \textbf{mAP}& \textbf{AUROC}&\textbf{mIOU}& \textbf{mAP} & \textbf{AUROC} \\
\cmidrule{2-7}
& \multicolumn{3}{c}{Val} & \multicolumn{3}{|c}{Test}\\
\midrule
XCLIP~\cite{ma2022x}  &35.54& 42.46 & 56.30 & 31.79 & 40.82& 58.71\\
RefAtomNet (XCLIP) & \textbf{38.59} & \textbf{47.40} & \textbf{61.20}& \textbf{36.61} & \textbf{48.59}& \textbf{66.47} \\
\midrule
BLIPv2~\cite{li2023BLIP}  & 32.99 & 52.13& 66.56& 32.75& 53.19& 69.92 \\
RefAtomNet (BLIPv2) & \textbf{38.22} & \textbf{55.98} & \textbf{69.73} & \textbf{36.42} & \textbf{57.52} & \textbf{73.95}\\
\midrule
\bottomrule
\end{tabular}}
\end{minipage}%
\end{table}
The reference sentences given by different users may differ in daily life scenarios. To test the generalizability of the model towards different referring styles, we invoke API (gpt-3.5-turbo) of ChatGPT~\cite{openai2022chatgpt} to rephrase the test set description two times and then deliver the averaged performance for RAVAR on the original val and test sets, and the two rephrased val and test sets, where we select $4$ most outperforming baselines from our benchmark to construct this ablation study, \ie, Singularity~\cite{lei2022revealing}, AskAnything~\cite{2023videochat}, XCLIP~\cite{ma2022x}, and BLIPv2~\cite{li2023BLIP}, as shown in Tab.~\ref{tab:rephrase}. \texttt{RefAtomNet} delivers the best performances. 
We find out that the test time reference rephrasing will cause performance decay for localization and less decay for the atomic video-based action recognition delivered by our model. 

\noindent\textbf{Generalizability to Different Visual Textual Backbones.}
Since we use the best-performing baseline BLIPv2~\cite{li2023BLIP} as the visual and textual encoder in our model, it would be interesting to see if the proposed architecture can generalize to different encoder backbones. We thereby demonstrate another ablation study in Tab.~\ref{tab:gen_tvr}, where we equip our \texttt{RefAtomNet} with the XCLIP~\cite{ma2022x} backbone. Compared with the XCLIP baseline, the \texttt{RefAtomNet} (XCLIP) achieves $3.05\%$, $4.94\%$, $4.90\%$ and $4.82\%$, $7.77\%$, $7.76\%$ performance improvements in terms of mIOU, mAP, and AUROC on the val and test sets, respectively, showing the great generalizability of our \texttt{RefAtomNet} of the visual and textual encoder.

\begin{figure*}[t!]
\centering
\includegraphics[width=1\linewidth]{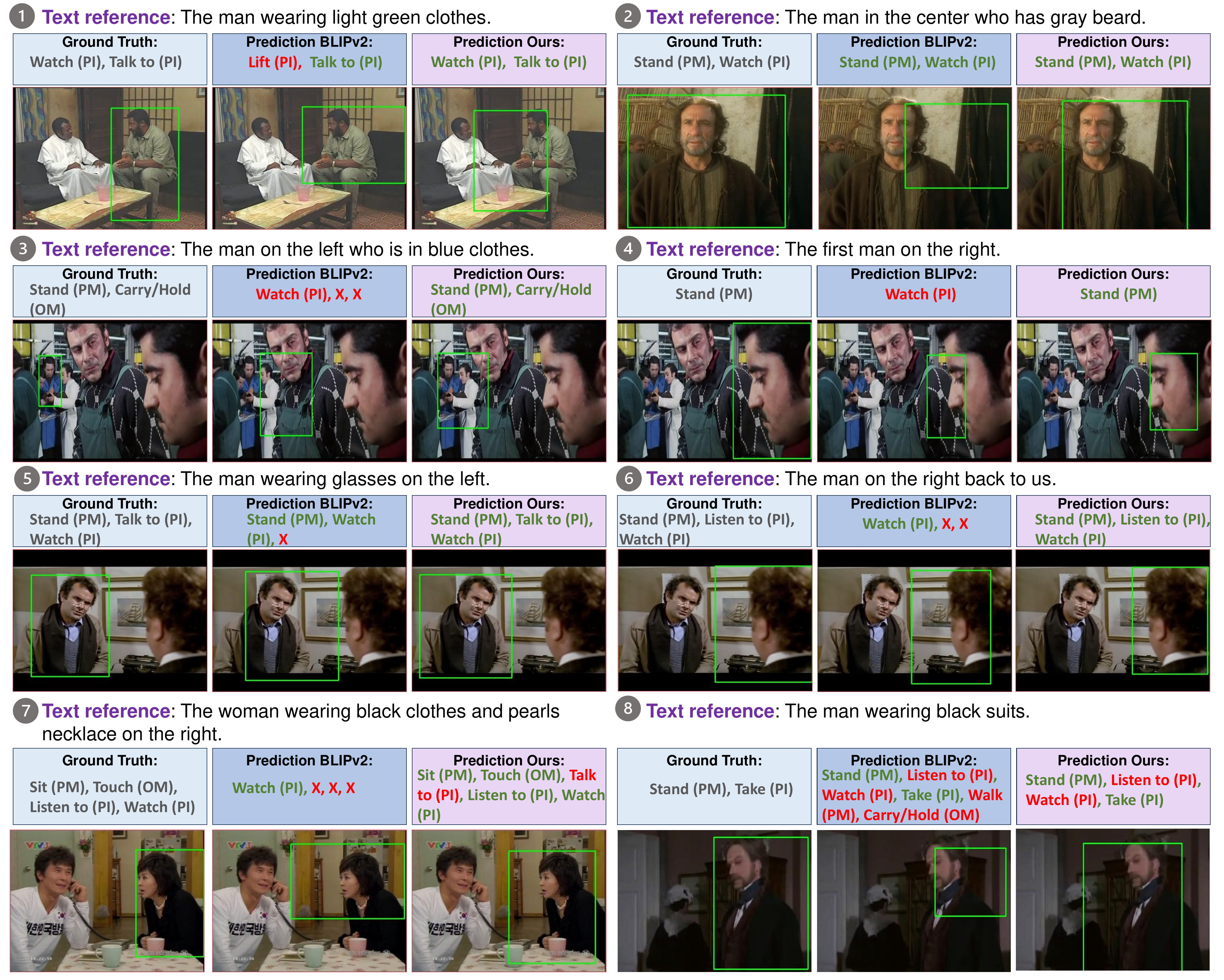}

\caption{An overview of qualitative results. Missed predictions are marked with a red cross, while true positive and false positive predictions are shown in green and red.}

\label{fig:probability_score1}
\end{figure*}
\noindent\textbf{Analysis of the Qualitative Results.}
The qualitative results are delivered in Fig.~\ref{fig:probability_score1}.
These examples demonstrate the effectiveness of \texttt{RefAtomNet} compared with the best-performing baseline BLIPv2~\cite{li2023BLIP}. Samples~\circlegray{1} and \circlegray{8} validate the recognition quality for references without any location indications. Samples \circlegray{3}, \circlegray{4}, \circlegray{5}, and \circlegray{6} showcase the results for different humans in a shared scene. Samples \circlegray{2} and \circlegray{7} show the RAVAR performance for references containing finer details, \eg, the necklace type and the beard color. 
Our \texttt{RefAtomNet} outperforms the BLIPv2 baseline on selected samples obviously, attributable to its superior capability in eliminating extraneous visual cues and its advanced location-semantic reasoning prowess.

\section{Conclusions}
In this work, we introduce a novel task called Referring Atomic Video Action Recognition (RAVAR).
We establish the RefAVA dataset to address the challenge of identifying atomic actions for individuals of interest in videos based on textual descriptions. Existing methods exhibit poor performance on this new task, prompting the development of \texttt{RefAtomNet}, a vision-language architecture that effectively integrates cross-stream tokens for precise referring atomic video action recognition. Through reference-relevant token enhancement and comprehensive token integration, \texttt{RefAtomNet} achieves impressive results, highlighting its effectiveness in tackling the complexities of the RAVAR task.
\section*{Acknowledgements}
The project served to prepare the SFB 1574 Circular Factory for the Perpetual Product (project ID: 471687386), approved by the German Research Foundation (DFG, German Research Foundation) with a start date of April 1, 2024. This work was also partially supported in part by the SmartAge project sponsored by the Carl Zeiss Stiftung (P2019-01-003; 2021-2026). This work was performed on the HoreKa supercomputer funded by the Ministry of Science, Research and the Arts Baden-Württemberg and by the Federal Ministry of Education and Research. The authors also acknowledge support by the state of Baden-Württemberg through bwHPC and the German Research Foundation (DFG) through grant INST 35/1597-1 FUGG. This project is also supported by the National Key RD Program under Grant 2022YFB4701400. Lastly, the authors thank for the support of Dr. Sepideh Pashami, the Swedish Innovation Agency VINNOVA, the Digital Futures.

\bibliographystyle{splncs04}
\bibliography{main}

\begin{thebibliography}{100}
\providecommand{\url}[1]{\texttt{#1}}
\providecommand{\urlprefix}{URL }
\providecommand{\doi}[1]{https://doi.org/#1}

\bibitem{bagad2023test}
Bagad, P., Tapaswi, M., Snoek, C.G.M.: Test of time: Instilling video-language models with a sense of time. In: CVPR (2023)

\bibitem{bu2022scene}
Bu, Y., Li, L., Xie, J., Liu, Q., Cai, Y., Huang, Q., Li, Q.: Scene-text oriented referring expression comprehension. TMM  (2022)

\bibitem{carion2020end}
Carion, N., Massa, F., Synnaeve, G., Usunier, N., Kirillov, A., Zagoruyko, S.: End-to-end object detection with transformers. In: ECCV (2020)

\bibitem{carreira2017quo}
Carreira, J., Zisserman, A.: Quo vadis, action recognition? {A} new model and the kinetics dataset. In: CVPR (2017)

\bibitem{castro2022wild}
Castro, S., Deng, N., Huang, P., Burzo, M., Mihalcea, R.: In-the-wild video question answering. In: COLING (2022)

\bibitem{chai2023stablevideo}
Chai, W., Guo, X., Wang, G., Lu, Y.: {StableVideo:} {Text-driven} consistency-aware diffusion video editing. In: ICCV (2023)

\bibitem{chen2023video}
Chen, J., Zhu, D., Haydarov, K., Li, X., Elhoseiny, M.: {Video ChatCaptioner:} {Towards} enriched spatiotemporal descriptions. arXiv preprint arXiv:2304.04227  (2023)

\bibitem{chen2015microsoft}
Chen, X., Fang, H., Lin, T.Y., Vedantam, R., Gupta, S., Doll{\'a}r, P., Zitnick, C.L.: Microsoft {COCO} captions: {Data} collection and evaluation server. arXiv preprint arXiv:1504.00325  (2015)

\bibitem{chen2023tagging}
Chen, Y., Wang, J., Lin, L., Qi, Z., Ma, J., Shan, Y.: Tagging before alignment: Integrating multi-modal tags for video-text retrieval. arXiv preprint arXiv:2301.12644  (2023)

\bibitem{chen2019weakly}
Chen, Z., Ma, L., Luo, W., Wong, K.Y.K.: Weakly-supervised spatio-temporally grounding natural sentence in video. arXiv preprint arXiv:1906.02549  (2019)

\bibitem{chung2021haa500}
Chung, J., Wuu, C.h., Yang, H.r., Tai, Y.W., Tang, C.K.: {HAA500:} {Human-centric} atomic action dataset with curated videos. In: ICCV (2021)

\bibitem{dang2023instructdet}
Dang, R., Feng, J., Zhang, H., Ge, C., Song, L., Gong, L., Liu, C., Chen, Q., Zhu, F., Zhao, R., Song, Y.: {InstructDET:} {Diversifying} referring object detection with generalized instructions. arXiv preprint arXiv:2310.05136  (2023)

\bibitem{deruyttere2019talk2car}
Deruyttere, T., Vandenhende, S., Grujicic, D., Van~Gool, L., Moens, M.F.: {Talk2Car:} {Taking} control of your self-driving car. In: EMNLP (2019)

\bibitem{Devlin2019BERTPO}
Devlin, J., Chang, M.W., Lee, K., Toutanova, K.: {BERT:} {Pre-training} of deep bidirectional transformers for language understanding. In: ACL (2019)

\bibitem{dosovitskiy2020vit}
Dosovitskiy, A., Beyer, L., Kolesnikov, A., Weissenborn, D., Zhai, X., Unterthiner, T., Dehghani, M., Minderer, M., Heigold, G., Gelly, S., Uszkoreit, J., Houlsby, N.: An image is worth 16x16 words: Transformers for image recognition at scale. In: ICLR (2021)

\bibitem{feichtenhofer2020x3d}
Feichtenhofer, C.: {X3D:} {Expanding} architectures for efficient video recognition. In: CVPR (2020)

\bibitem{feichtenhofer2019slowfast}
Feichtenhofer, C., Fan, H., Malik, J., He, K.: {SlowFast} networks for video recognition. In: ICCV (2019)

\bibitem{gandhi2022measuring}
Gandhi, M., Gul, M.O., Prakash, E., Grunde-McLaughlin, M., Krishna, R., Agrawala, M.: Measuring compositional consistency for video question answering. In: CVPR (2022)

\bibitem{gao2023mist}
Gao, D., Zhou, L., Ji, L., Zhu, L., Yang, Y., Shou, M.Z.: {MIST:} {Multi-modal} iterative spatial-temporal transformer for long-form video question answering. In: CVPR (2023)

\bibitem{garcia2020knowit}
Garcia, N., Otani, M., Chu, C., Nakashima, Y.: {KnowIT VQA:} {Answering} knowledge-based questions about videos. In: AAAI (2020)

\bibitem{gavrilyuk2018actor}
Gavrilyuk, K., Ghodrati, A., Li, Z., Snoek, C.G.: Actor and action video segmentation from a sentence. In: CVPR (2018)

\bibitem{goyal2017something}
Goyal, R., Kahou, S.E., Michalski, V., Materzynska, J., Westphal, S., Kim, H., Haenel, V., Fr{\"{u}}nd, I., Yianilos, P., Mueller{-}Freitag, M., Hoppe, F., Thurau, C., Bax, I., Memisevic, R.: The ``something something'' video database for learning and evaluating visual common sense. In: ICCV (2017)

\bibitem{gritsenko2023end}
Gritsenko, A., Xiong, X., Djolonga, J., Dehghani, M., Sun, C., Lu{\v{c}}i{\'c}, M., Schmid, C., Arnab, A.: End-to-end spatio-temporal action localisation with video transformers. arXiv preprint arXiv:2304.12160  (2023)

\bibitem{gu2018ava}
Gu, C., Sun, C., Ross, D.A., Vondrick, C., Pantofaru, C., Li, Y., Vijayanarasimhan, S., Toderici, G., Ricco, S., Sukthankar, R., Schmid, C., Malik, J.: {AVA:} {A} video dataset of spatio-temporally localized atomic visual actions. In: CVPR (2018)

\bibitem{guo2021re}
Guo, W., Zhang, Y., Yang, J., Yuan, X.: Re-attention for visual question answering. TIP  (2021)

\bibitem{han2023agent}
Han, D., Ye, T., Han, Y., Xia, Z., Song, S., Huang, G.: Agent attention: On the integration of softmax and linear attention. arXiv preprint arXiv:2312.08874  (2023)

\bibitem{ji2019context}
Ji, Y., Zhan, Y., Yang, Y., Xu, X., Shen, F., Shen, H.T.: A context knowledge map guided coarse-to-fine action recognition. TIP  (2020)

\bibitem{jiang2020divide}
Jiang, J., Chen, Z., Lin, H., Zhao, X., Gao, Y.: Divide and conquer: Question-guided spatio-temporal contextual attention for video question answering. In: AAAI (2020)

\bibitem{jin2023refclip}
Jin, L., Luo, G., Zhou, Y., Sun, X., Jiang, G., Shu, A., Ji, R.: Refclip: A universal teacher for weakly supervised referring expression comprehension. In: CVPR (2023)

\bibitem{khoreva2019video}
Khoreva, A., Rohrbach, A., Schiele, B.: Video object segmentation with language referring expressions. In: ACCV (2019)

\bibitem{kim2024atrous}
Kim, M., Spinola, F., Benz, P., Kim, T.h.: A*: Atrous spatial temporal action recognition for real time applications. In: WACV (2024)

\bibitem{kingma2014adam}
Kingma, D.P., Ba, J.: Adam: A method for stochastic optimization. arXiv preprint arXiv:1412.6980  (2014)

\bibitem{kirillov2023segment}
Kirillov, A., Mintun, E., Ravi, N., Mao, H., Rolland, C., Gustafson, L., Xiao, T., Whitehead, S., Berg, A.C., Lo, W.Y., et~al.: Segment anything. In: CVPR (2023)

\bibitem{kuehne2011hmdb}
Kuehne, H., Jhuang, H., Garrote, E., Poggio, T., Serre, T.: {HMDB:} {A} large video database for human motion recognition. In: ICCV (2011)

\bibitem{laput2019sensing}
Laput, G., Harrison, C.: Sensing fine-grained hand activity with smartwatches. In: CHI (2019)

\bibitem{le2020hierarchical}
Le, T.M., Le, V., Venkatesh, S., Tran, T.: Hierarchical conditional relation networks for video question answering. In: CVPR (2020)

\bibitem{lea2016learning}
Lea, C., Vidal, R., Hager, G.D.: Learning convolutional action primitives for fine-grained action recognition. In: ICRA (2016)

\bibitem{lei2022revealing}
Lei, J., Berg, T.L., Bansal, M.: Revealing single frame bias for video-and-language learning. arXiv preprint arXiv:2206.03428  (2022)

\bibitem{lei2021less}
Lei, J., Li, L., Zhou, L., Gan, Z., Berg, T.L., Bansal, M., Liu, J.: Less is more: {ClipBERT} for video-and-language learning via sparse sampling. In: CVPR (2021)

\bibitem{li2022learning}
Li, G., Wei, Y., Tian, Y., Xu, C., Wen, J.R., Hu, D.: Learning to answer questions in dynamic audio-visual scenarios. In: CVPR (2022)

\bibitem{li2022representation}
Li, J., Niu, L., Zhang, L.: From representation to reasoning: Towards both evidence and commonsense reasoning for video question-answering. In: CVPR (2022)

\bibitem{li2023BLIP}
Li, J., Li, D., Savarese, S., Hoi, S.: {BLIP-2:} {Bootstrapping} language-image pre-training with frozen image encoders and large language models. In: ICML (2023)

\bibitem{li2022BLIP}
Li, J., Li, D., Xiong, C., Hoi, S.: {BLIP:} {Bootstrapping} language-image pre-training for unified vision-language understanding and generation. In: ICML (2022)

\bibitem{2023videochat}
Li, K., He, Y., Wang, Y., Li, Y., Wang, W., Luo, P., Wang, Y., Wang, L., Qiao, Y.: {VideoChat:} {Chat-centric} video understanding. arXiv preprint arXiv:2305.06355  (2023)

\bibitem{li2023lavender}
Li, L., Gan, Z., Lin, K., Lin, C.C., Liu, Z., Liu, C., Wang, L.: {LAVENDER:} {Unifying} video-language understanding as masked language modeling. In: CVPR (2023)

\bibitem{li2018referring}
Li, R., Li, K., Kuo, Y.C., Shu, M., Qi, X., Shen, X., Jia, J.: Referring image segmentation via recurrent refinement networks. In: CVPR (2018)

\bibitem{li2022mvitv2}
Li, Y., Wu, C.Y., Fan, H., Mangalam, K., Xiong, B., Malik, J., Feichtenhofer, C.: {MViTv2:} {Improved} multiscale vision transformers for classification and detection. In: CVPR (2022)

\bibitem{lin2024echotrack}
Lin, J., Chen, J., Peng, K., He, X., Li, Z., Stiefelhagen, R., Yang, K.: {EchoTrack:} {Auditory} referring multi-object tracking for autonomous driving. arXiv preprint arXiv:2402.18302  (2024)

\bibitem{lin2017focal}
Lin, T.Y., Goyal, P., Girshick, R., He, K., Doll{\'a}r, P.: Focal loss for dense object detection. In: ICCV (2017)

\bibitem{lin2023towards}
Lin, X., Tiwari, S., Huang, S., Li, M., Shou, M.Z., Ji, H., Chang, S.F.: Towards fast adaptation of pretrained contrastive models for multi-channel video-language retrieval. In: CVPR (2023)

\bibitem{liu2017referring}
Liu, J., Wang, L., Yang, M.H.: Referring expression generation and comprehension via attributes. In: ICCV (2017)

\bibitem{liu2023open}
Liu, R., Zhang, J., Peng, K., Zheng, J., Cao, K., Chen, Y., Yang, K., Stiefelhagen, R.: Open scene understanding: Grounded situation recognition meets segment anything for helping people with visual impairments. In: ICCVW (2023)

\bibitem{liu2019clevr}
Liu, R., Liu, C., Bai, Y., Yuille, A.L.: {CLEVR-Ref+:} {Diagnosing} visual reasoning with referring expressions. In: CVPR (2019)

\bibitem{liu2021cross}
Liu, S., Hui, T., Huang, S., Wei, Y., Li, B., Li, G.: Cross-modal progressive comprehension for referring segmentation. TPAMI  (2021)

\bibitem{liu2023cross}
Liu, Y., Li, G., Lin, L.: Cross-modal causal relational reasoning for event-level visual question answering. TPAMI  (2023)

\bibitem{luo2022clip4clip}
Luo, H., Ji, L., Zhong, M., Chen, Y., Lei, W., Duan, N., Li, T.: {CLIP4Clip:} {An} empirical study of {CLIP} for end to end video clip retrieval and captioning. Neurocomputing  (2022)

\bibitem{ma2022x}
Ma, Y., Xu, G., Sun, X., Yan, M., Zhang, J., Ji, R.: {X-CLIP:} {End-to-end} multi-grained contrastive learning for video-text retrieval. In: MM (2022)

\bibitem{madasu2023improving}
Madasu, A., Aflalo, E., Ben Melech~Stan, G., Tseng, S.Y., Bertasius, G., Lal, V.: Improving video retrieval using multilingual knowledge transfer. In: ECIR (2023)

\bibitem{mcintosh2020visual}
McIntosh, B., Duarte, K., Rawat, Y.S., Shah, M.: Visual-textual capsule routing for text-based video segmentation. In: CVPR (2020)

\bibitem{nagrani2021attention}
Nagrani, A., Yang, S., Arnab, A., Jansen, A., Schmid, C., Sun, C.: Attention bottlenecks for multimodal fusion. In: NeuIPS (2021)

\bibitem{openai2022chatgpt}
OpenAI: {ChatGPT:} {Optimizing} language models for dialogue. \url{https://openai.com/} (2022)

\bibitem{ordonez2011im2text}
Ordonez, V., Kulkarni, G., Berg, T.: {Im2Text:} {Describing} images using 1 million captioned photographs. In: NeurIPS (2011)

\bibitem{ou2022indoor}
Ou, W., Zhang, J., Peng, K., Yang, K., Jaworek, G., M{\"u}ller, K., Stiefelhagen, R.: Indoor navigation assistance for visually impaired people via dynamic {SLAM} and panoptic segmentation with an {RGB-D} sensor. In: ICCHP (2022)

\bibitem{peng2022transdarc}
Peng, K., Roitberg, A., Yang, K., Zhang, J., Stiefelhagen, R.: {TransDARC:} {Transformer-based} driver activity recognition with latent space feature calibration. In: IROS (2022)

\bibitem{pramanick2022doro}
Pramanick, P., Sarkar, C., Paul, S., dev Roychoudhury, R., Bhowmick, B.: {DoRO:} {Disambiguation} of referred object for embodied agents. RA-L  (2022)

\bibitem{pramono2021spatial}
Pramono, R.R.A., Chen, Y.T., Fang, W.H.: Spatial-temporal action localization with hierarchical self-attention. TMM  (2021)

\bibitem{qiu2020language}
Qiu, H., Li, H., Wu, Q., Meng, F., Shi, H., Zhao, T., Ngan, K.N.: Language-aware fine-grained object representation for referring expression comprehension. In: MM (2020)

\bibitem{radford2021learning}
Radford, A., Kim, J.W., Hallacy, C., Ramesh, A., Goh, G., Agarwal, S., Sastry, G., Askell, A., Mishkin, P., Clark, J., Krueger, G., Sutskever, I.: Learning transferable visual models from natural language supervision. In: ICML (2021)

\bibitem{rajasegaran2023benefits}
Rajasegaran, J., Pavlakos, G., Kanazawa, A., Feichtenhofer, C., Malik, J.: On the benefits of {3D} pose and tracking for human action recognition. In: CVPR (2023)

\bibitem{ryali2023hiera}
Ryali, C., Hu, Y., Bolya, D., Wei, C., Fan, H., Huang, P., Aggarwal, V., Chowdhury, A., Poursaeed, O., Hoffman, J., Malik, J., Li, Y., Feichtenhofer, C.: Hiera: A hierarchical vision transformer without the bells-and-whistles. In: ICML (2023)

\bibitem{saha2018fine}
Saha, J., Chowdhury, C., Chowdury, I.R., Roy, P.: Fine grained activity recognition using smart handheld. In: ICDCN (2018)

\bibitem{seibold2022reference}
Seibold, C.M., Rei{\ss}, S., Kleesiek, J., Stiefelhagen, R.: Reference-guided pseudo-label generation for medical semantic segmentation. In: AAAI (2022)

\bibitem{seo2020urvos}
Seo, S., Lee, J.Y., Han, B.: {URVOS:} {Unified} referring video object segmentation network with a large-scale benchmark. In: ECCV (2020)

\bibitem{shao2020finegym}
Shao, D., Zhao, Y., Dai, B., Lin, D.: Finegym: A hierarchical video dataset for fine-grained action understanding. In: CVPR (2020)

\bibitem{sharma2018conceptual}
Sharma, P., Ding, N., Goodman, S., Soricut, R.: Conceptual captions: A cleaned, hypernymed, image alt-text dataset for automatic image captioning. In: ACL (2018)

\bibitem{shi2023unsupervised}
Shi, H., Pan, W., Zhao, Z., Zhang, M., Wu, F.: Unsupervised domain adaptation for referring semantic segmentation. In: MM (2023)

\bibitem{shi2018key}
Shi, H., Li, H., Meng, F., Wu, Q.: Key-word-aware network for referring expression image segmentation. In: ECCV (2018)

\bibitem{shi2023learning}
Shi, Y., Xu, H., Yuan, C., Li, B., Hu, W., Zha, Z.J.: Learning video-text aligned representations for video captioning. TOMM  (2023)

\bibitem{soomro2012ucf101}
Soomro, K., Zamir, A.R., Shah, M.: {UCF101:} {A} dataset of 101 human actions classes from videos in the wild. arXiv preprint arXiv:1212.0402  (2012)

\bibitem{su2023sequence}
Su, Y., Wang, W., Liu, J., Ma, S., Yang, X.: Sequence as a whole: A unified framework for video action localization with long-range text query. TIP  (2023)

\bibitem{vasudevan2018object}
Vasudevan, A.B., Dai, D., Van~Gool, L.: Object referring in videos with language and human gaze. In: CVPR (2018)

\bibitem{wang2023videomae}
Wang, L., Huang, B., Zhao, Z., Tong, Z., He, Y., Wang, Y., Wang, Y., Qiao, Y.: {VideoMAE V2:} {Scaling} video masked autoencoders with dual masking. In: CVPR (2023)

\bibitem{wang2023actionclip}
Wang, M., Xing, J., Mei, J., Liu, Y., Jiang, Y.: {ActionCLIP:} {Adapting} language-image pretrained models for video action recognition. TNNLS  (2023)

\bibitem{wang2023masked}
Wang, R., Chen, D., Wu, Z., Chen, Y., Dai, X., Liu, M., Yuan, L., Jiang, Y.G.: Masked video distillation: Rethinking masked feature modeling for self-supervised video representation learning. In: CVPR (2023)

\bibitem{wang2023stal}
Wang, S., Yan, R., Huang, P., Dai, G., Song, Y., Shu, X.: {Com-STAL:} {Compositional} spatio-temporal action localization. TCSVT  (2023)

\bibitem{wang2022internvideo}
Wang, Y., Li, K., Li, Y., He, Y., Huang, B., Zhao, Z., Zhang, H., Xu, J., Liu, Y., Wang, Z., Xing, S., Chen, G., Pan, J., Yu, J., Wang, Y., Wang, L., Qiao, Y.: {InternVideo:} {General} video foundation models via generative and discriminative learning. arXiv preprint arXiv:2212.03191  (2022)

\bibitem{wu2023referring}
Wu, D., Han, W., Wang, T., Dong, X., Zhang, X., Shen, J.: Referring multi-object tracking. In: CVPR (2023)

\bibitem{wu2023cap4video}
Wu, W., Luo, H., Fang, B., Wang, J., Ouyang, W.: {Cap4Video:} {What} can auxiliary captions do for text-video retrieval? In: CVPR (2023)

\bibitem{xiao2021next}
Xiao, J., Shang, X., Yao, A., Chua, T.S.: {NExT-QA:} {Next} phase of question-answering to explaining temporal actions. In: CVPR (2021)

\bibitem{yang2021just}
Yang, A., Miech, A., Sivic, J., Laptev, I., Schmid, C.: Just ask: Learning to answer questions from millions of narrated videos. In: ICCV (2021)

\bibitem{yang2022avqa}
Yang, P., Wang, X., Duan, X., Chen, H., Hou, R., Jin, C., Zhu, W.: {AVQA:} {A} dataset for audio-visual question answering on videos. In: MM (2022)

\bibitem{Yi2021BenchmarkingTR}
Yi, C., Yang, S., Li, H., Tan, Y.P., Kot, A.C.: Benchmarking the robustness of spatial-temporal models against corruptions. In: NeurIPS (2021)

\bibitem{yu2016modeling}
Yu, L., Poirson, P., Yang, S., Berg, A.C., Berg, T.L.: Modeling context in referring expressions. In: ECCV (2016)

\bibitem{yu2019activitynet}
Yu, Z., Xu, D., Yu, J., Yu, T., Zhao, Z., Zhuang, Y., Tao, D.: {ActivityNet-QA:} {A} dataset for understanding complex web videos via question answering. In: AAAI (2019)

\bibitem{yuan2021instancerefer}
Yuan, Z., Yan, X., Liao, Y., Zhang, R., Wang, S., Li, Z., Cui, S.: {InstanceRefer:} {Cooperative} holistic understanding for visual grounding on point clouds through instance multi-level contextual referring. In: ICCV (2021)

\bibitem{zeng2022motr}
Zeng, F., Dong, B., Zhang, Y., Wang, T., Zhang, X., Wei, Y.: {MOTR:} {End-to-end} multiple-object tracking with transformer. In: ECCV (2022)

\bibitem{zhang2023multi}
Zhang, G., Ren, J., Gu, J., Tresp, V.: Multi-event video-text retrieval. In: CVPR (2023)

\bibitem{zhang2023video}
Zhang, H., Li, X., Bing, L.: {Video-LLaMA:} {An} instruction-tuned audio-visual language model for video understanding. In: EMNLP (2023)

\bibitem{zheng2023materobot}
Zheng, J., Zhang, J., Yang, K., Peng, K., Stiefelhagen, R.: {MateRobot:} {Material} recognition in wearable robotics for people with visual impairments. In: ICRA (2024)

\bibitem{zong2023mcomet}
Zong, D., Sun, S.: {McOmet:} {Multimodal} fusion transformer for physical audiovisual commonsense reasoning. In: AAAI (2023)

\end{thebibliography}

\appendix
\section{Potential Impacts}
In this section, we deliberate on the societal implications of our research endeavors. We introduce a new task, Referring Atomic Video Action Recognition (RAVAR), by developing the RefAVA dataset and establishing the RAVAR benchmark. A total of $36,630$ instances were meticulously annotated based on videos sourced from the AVA dataset to facilitate subsequent inquiries into the RAVAR domain. In contrast to traditional approaches in atomic video action detection, which typically depend on post-hoc selections of the person of interest, our methodology leverages textual descriptions as indicators. This approach mitigates the reliance on precise positional data delineated by bounding boxes, which may exhibit significant variability over time.

The introduction of this task setting is crucial for advancing deep learning models' capabilities in comprehending scenes with greater details, particularly within the atomic video action recognition framework.
Such enhanced understanding is challenging and important in scenarios involving complex interactions among multiple individuals, as is often encountered in fields such as rehabilitation and robotic assistance.
This shift towards a more nuanced understanding has the potential to significantly impact these applications, offering more adaptable and context-aware solutions.

To construct the first testbed for the RAVAR task, we use $11$ well-established methods from Atomic Action Localization (AAL), Text Video Retrieval (TVR), and Video Question Answering (VQA) domains while observing that most of the selected baselines cannot deliver satisfactory performances. AAL approaches are not good at dealing with the referring location of the person of interest while TVR and VQA approaches are not good at handling the fine-grained action recognition. We propose \texttt{RefAtomNet} to address both of the aforementioned issues. 
\texttt{RefAtomNet} relies on three streams of token extraction, namely the textual reference tokens, the visual tokens, and the newly proposed location-semantic tokens, which are used for incorporating semantic and location cues provided by well-established object detectors from the scene. 
\texttt{RefAtomNet} further takes advantage of the 1D sequential agent attention on each stream of the tokens to achieve self-suppression regarding the irrelevant tokens. It then utilizes the agent-based location semantic aware attentional fusion to enhance the visual stream by merging the agent attention masks and the agent tokens from the textual reference and the location-semantic streams. 
The proposed \texttt{RefAtomNet} achieves state-of-the-art on the RAVAR benchmark, with great generalizability towards the visual textual backbone and object detector. \texttt{RefAtomNet} can also deliver promising RAVAR performances when faced with the test time reference rephrasing and test time video disturbances compared with the chosen baselines, which benefits practical deployment. However, our method still has the potential to output false predictions and biased content, which can have undesired consequences, impacting society negatively.

\section{Discussion of Limitations} 
RAVAR by referring to multiple persons preserving the same attributes though one sentence in one video is not tackled in our work, since we annotate each individual and ensure that the described person can be successfully referred by the provided description, and there are no two persons who share the same reference sentence in one video clip. However, we regard this direction as an interesting future work direction for the RAVAR task. Since we deliver the first dataset for the RAVAR field and there is no other existing dataset for this new task, our experiments are only conducted on the contributed RefAVA dataset.

\section{Evaluation Metrics}
\label{sec-3-3}

\subsection{Mean Average Precision (mAP)} 
The mAP for multi-label classification measures prediction precision across labels by computing precision at different thresholds and plotting precision-recall curves for each label. 
The Average Precision (AP) for each label is derived by integrating over these curves, typically approximated by summing areas under specific points. The mAP, an average of these AP values across all labels, serves as an aggregate performance metric for accurately predicting multiple labels per instance. This aggregate performance measure reflects the model's proficiency in accurately predicting multiple labels for each instance within the dataset. It accounts for the model's ability not only to identify the presence of various labels but also to ascertain their absence, thereby ensuring a balanced evaluation of its predictive capabilities across all possible label outcomes.

\subsection{Area Under the Receiver Operating Characteristic (AUROC)} 
The AUROC for multi-label classification is determined by considering each label as a separate binary classification and calculating the True Positive Rate (TPR) and False Positive Rate (FPR) to plot ROC curves for each label.
The overall AUROC for the multi-label classification model is then calculated by averaging the individual AUROC scores across all labels. This aggregated metric, often referred to as the macro-average AUROC, provides a global indicator of the model’s discriminative capability across the entire multi-label task. It is a critical measure, particularly where the balance between different classes varies significantly, as it offers an unbiased metric that does not favor labels with more instances.
Emphasizing the importance of AUROC in multi-label classification highlights its role in ensuring the model’s robustness and effectiveness across diverse conditions. 

\subsection{Mean Intersection of Union (mIOU)} 
The mIOU is a prominent evaluation measure used in the context of bounding box regression tasks, particularly in object detection. We use mIOU to evaluate the bounding box regression ability for the person of interest. 
This metric is chosen as an auxiliary indicator that has less priority than the mAP and the AUROC metrics since our main aim is to harvest the correct atomic action predictions for the referring person. 

\subsection{Further Clarification regarding the Metrics}
Our task prioritizes recognition over localization to progressively address this new RAVAR challenge. We evaluate detection and recognition performance separately, treating mIOU as a secondary metric. We provide mAP (IOU=0.2, 0.5) of our approach and BLIPv2, which delivers the best performances among all leveraged baselines, as follows.

\begin{table}[h]
    \centering
    \scalebox{1}{
    \begin{tabular}{c|cc|cc}
        \toprule
       \midrule
        & \textbf{BLIPv2-Val} & \textbf{BLIPv2-Test} & \textbf{Ours-Val} & \textbf{Ours-Test} \\
       \hline
       mAP (IOU=0.2) & 46.77 & 45.21 & \textbf{51.26} & \textbf{49.47}  \\
       mAP (IOU=0.5) & 41.40 & 39.27  & \textbf{44.84} & \textbf{41.93} \\
\midrule
\bottomrule
\end{tabular}
    }
    \label{tab:supplement_metrics}
\end{table}

\section{Generalizability to Different Detectors}
In this work, we also deliver the ablation towards how much will the object detector affect the performance of RAVAR in Tab.~\ref{tab:detector}. We conduct an ablation study by replacing the DETR~\cite{carion2020end}, which serves as the object detector in our \texttt{RefAtomNet}, by using RetinaNet~\cite{lin2017focal}. Compared with the most outperforming baselines BLIPv2~\cite{li2023BLIP}, both \texttt{RefAtomNet} (RetinaNet) and \texttt{RefAtomNet} (DETR) show promising performance improvements. Regarding the two employed object detectors, there are slight performance changes with $< 1\%$ for each metric, showcasing that our proposed \texttt{RefAtomNet} can generalize well to other object detectors. \texttt{RefAtomNet} (RetinaNet) can harvest $38.65\%$, $55.08\%$, and $69.16\%$ of mIOU, mAP, and AUROC and $37.82\%$, $56.67\%$, and $73.63\%$ of mIOU, mAP, and AUROC, on val and test sets, respectively.
\begin{table}[t]
\caption{Generalizability to other object detectors}
\label{tab:detector}
\centering
\begin{tabular}{l|ccc|ccc}
\toprule
\midrule
\multirow{2}{*}{\textbf{Fusion}} & \textbf{mIOU}   & \textbf{mAP}& \textbf{AUROC}&\textbf{mIOU}& \textbf{mAP} & \textbf{AUROC} \\
\cmidrule{2-7}
& \multicolumn{3}{c}{Val} & \multicolumn{3}{|c}{Test}\\
\midrule
BLIPv2~\cite{li2023BLIP} & 32.99 & 52.13& 66.56& 32.75& 53.19& 69.92 \\
RefAtomNet (RetinaNet~\cite{lin2017focal})  & \textbf{38.65} & 55.08 &69.16& \textbf{37.82} & 56.67 &73.63 \\
RefATomNet (DETR~\cite{carion2020end}) & 38.22 & \textbf{55.98} & \textbf{69.73} & 36.42 & \textbf{57.52} & \textbf{73.95}\\
\midrule
\bottomrule
\end{tabular}
\end{table}

\section{Ablation of the Module Parameters}
In this section, we deliver the analysis on the hyperparameters of the proposed \texttt{RefAtomNet} by using BLIPv2~\cite{li2023BLIP} as the textual visual backbone and DETR~\cite{carion2020end} as the object detector to pursue the suitable hyperparameters for the number of heads and the number of agents.
\subsection{Ablation of the Frame Number}
Overall we get better performance when using $8$ frames in training and testing, which is the frame number we used in our main paper.
\begin{table}[h]
\caption{Ablation of the frame number.}
\centering
\begin{tabular}{l|ccc|ccc}
\toprule
\midrule
\multirow{2}{*}{\textbf{Frames}} & \textbf{mIOU}   & \textbf{mAP}& \textbf{AUROC}&\textbf{mIOU}& \textbf{mAP} & \textbf{AUROC} \\
\cmidrule{2-7}
& \multicolumn{3}{c}{Val} & \multicolumn{3}{|c}{Test}\\
\midrule
4  & 37.82 & 53.21 & 67.27 & 34.75 & 55.00 & 72.23 \\
6 & 37.28 & 53.55 & 67.50 & 36.01 & 55.24 & 72.14 \\
\textbf{8} & \textbf{38.22} & \textbf{55.98} & \textbf{69.73} & \textbf{36.42} & \textbf{57.52} & \textbf{73.95} \\
10 & 37.51 & 53.41 & 67.33 & 36.12 & 54.87 & 71.86 \\
12 & 36.85 & 53.37 & 67.36 & 36.22 & 55.20 & 71.90 \\
\midrule
\bottomrule
\end{tabular}
\label{tab:frame}
\end{table}
\subsection{Ablation of the Head Number}
We deliver the ablation of the head number used for acquiring the Query, Key, Value, and Agent in Tab.~\ref{tab:head}, where head number $N_h \in \left[1,2,3,4,16\right]$. We observe that when $N_h=1$, the \texttt{RefAtomNet} achieves the best performance. We thereby use $N_h=1$ in our \texttt{RefAtomNet}. All the experiments are conducted by selecting the number of agents $N_a=4$.
\begin{table}[t]
\caption{Ablation of the head number when the agent number is set as $4$.}
\label{tab:head}
\centering
\begin{tabular}{l|ccc|ccc}
\toprule
\midrule
\multirow{2}{*}{\textbf{Heads}} & \textbf{mIOU}   & \textbf{mAP}& \textbf{AUROC}&\textbf{mIOU}& \textbf{mAP} & \textbf{AUROC} \\
\cmidrule{2-7}
& \multicolumn{3}{c}{Val} & \multicolumn{3}{|c}{Test}\\
\midrule
1  & 38.22 & 55.98 & 69.73 & 36.42 & 57.52 & 73.95\\
2  & 35.60& 54.91 & 68.94& 34.72 & 55.02 & 71.71\\
3  &34.58 & 54.97 & 71.71& 34.58 & 54.97 & 71.71\\
4  &37.17 & 54.70 & 68.79 & 35.63 & 54.88 & 71.59\\
16  &36.10  &54.40  &68.59 & 35.22 & 54.68 & 71.40\\
\midrule
\bottomrule
\end{tabular}
\end{table}

\subsection{Ablation of the Agent Number} 
We further show the ablation study of the agent number in Tab.~\ref{tab:agent}, where $N_a \in \left[1,2,3,4,16\right]$. We observe that when $N_a{=}4$ the \texttt{RefAtomNet} achieves the best performance on val set considering the primary evaluation metrics, \ie, mAP and AUROC. We thereby use $N_a{=}4$ in our network setting.
\begin{table}[t]
\caption{Ablation of Agent Number when the head number is set as $1$.}
\label{tab:agent}
\centering
\begin{tabular}{l|ccc|ccc}
\toprule
\midrule
\multirow{2}{*}{\textbf{Agents}} & \textbf{mIOU}   & \textbf{mAP}& \textbf{AUROC}&\textbf{mIOU}& \textbf{mAP} & \textbf{AUROC} \\
\cmidrule{2-7}
& \multicolumn{3}{c}{Val} & \multicolumn{3}{|c}{Test}\\
\midrule
1  & 32.63 & 56.71 & 69.70& 33.68 & 55.40 &73.08\\
2  & 37.97 & 55.68 & 69.58 & 36.89 & 57.81 & 74.29 \\
3  &36.80  &55.86  & 69.82 & 35.37 & 57.32 & 73.77 \\
4  & 38.22 & 55.98 & 69.73 & 36.42 & 57.52 & 73.95\\
16  & 35.47 & 55.52 & 69.57& 33.96 & 57.91  & 74.22 \\
\midrule
\bottomrule
\end{tabular}
\end{table}

\section{Discussion of the Model Parameters}
\begin{table}[t]
\caption{
A comparison of the model trainable parameters and the performance. The mAP on the val and test sets are shown.}
\label{tab:parameters}

\scalebox{0.65}{\begin{tabular}{l|llllllllllll}
\toprule
\midrule
\textbf{Dataset} &I3D & X3D&MViTv2-B & VideoMAE2-B  & Hiera-B & Singularity & AskAnything & MeVTR & Clip4Clip & XClip&BLIPv2&RefAtomNet    \\
\midrule
\textbf{N$_{Parameters}$}  & 25M & 3.76M& 71M &87M &52M & 203M&0.66M &164M &149M &150M &187M&214M\\
\textbf{mAP$_{Val}$} & 44.04 & 44.45& 42.32& 42.02 & 42.74& 42.43& 51.42 &38.42  &39.48 &42.46 &52.13&55.98\\
\textbf{mAP$_{Test}$} & 44.64 & 46.34 &42.60 & 41.87  &41.14 & 42.18  &52.25 &36.27 &37.17 &40.82 &53.19 & 57.52\\
\midrule
\bottomrule
\end{tabular}}
\end{table}
We compare the baselines and our proposed method on the prioritized performances in terms of mAP for val and test sets, and the amount of the trainable parameters in Tab.~\ref{tab:parameters}. Most of the approaches used for the visual language model preserve more than $100M$ trainable parameters compared to the approaches from the atomic action localization group. Compared with the most outperforming baseline BLIPv2~\cite{li2023BLIP}, our proposed new modules only result in the increment of $27M$ parameters, while delivering promising mAP improvements by $3.85\%$ and $4.33\%$, on the val and test sets. Compared with the method of the largest scale in the baselines, \ie, Singularity, our method delivers $13.55\%$ and $15.34\%$ mAP benefits on the val and test sets with only $11M$ more parameters, indicating the effectiveness of our method for RAVAR by suppressing the irrelevant information in the visual stream.

\section{Robustness against Disturbances on Video during Test Time}

During practical usage, the input video has the possibility to be disturbed by different video noises. To assess the robustness of the proposed model and the most outperforming baselines, we conduct a robustness ablation study in Tab.~\ref{tab:robust_1} to simulate the rain and fog noises and Tab.~\ref{tab:robust_2} for shot and Gaussian noises on the val and test sets according to several perturbation types derived from~\cite{Yi2021BenchmarkingTR}. In the following, we will deliver the definition of different test time perturbations and the analysis of each perturbation in detail.

\subsection{Rain Noise}
The process of injecting rain noise into video frames is engineered to emulate the visual manifestation of precipitation within video sequences. This procedure can be delineated through the following steps:
\begin{table}[t]
\centering
\caption{Experimental results of the most outperforming baselines and \texttt{RefAtomNet} when rain noise and fog noise perturbations are added into the videos in the test phase.}
\label{tab:robust_1}
\scalebox{0.8}{\begin{tabular}{l|ccc|ccc|ccc|ccc}

\toprule
\midrule
\multirow{3}{*}{\textbf{Method}} & \multicolumn{6}{c|}{\textbf{Test-time rain noise perturbation}} & \multicolumn{6}{c}{\textbf{Test-time fog noise perturbation}} \\
\cmidrule{2-13}
& mIOU    & mAP    & AUROC   & mIOU   & mAP   & AUROC   & mIOU    & mAP    & AUROC   & mIOU   & mAP   & AUROC   \\
\cmidrule{2-13}
& \multicolumn{3}{c|}{Val}& \multicolumn{3}{c|}{Test} & \multicolumn{3}{c|}{Val} & \multicolumn{3}{c}{Test}\\
\midrule
Singularity~\cite{lei2021less} & 14.44& 39.32 & 56.17 & 15.35&  39.95& 53.28 &14.05  &39.46 & 56.68 &15.84 &40.58 & 53.43\\
XCLIP~\cite{ma2022x} &36.92 & 41.85 &55.53  &33.10 & 38.93 &57.15  & 37.24 &40.44 &52.90  &33.99 &36.33 & 52.41\\
AskAnything~\cite{2023videochat} &20.63 &51.05 & 65.69 & 21.91& 51.69 & 68.54 & 24.37 &49.56 & 63.51 &25.27 & 50.35& 66.78\\
BLIPv2~\cite{li2023BLIP} &31.53 & 51.26& 65.60 &31.93 & 53.15 & 69.55 & 35.85 &48.53& 63.18 & 35.65& 52.55& 68.00 \\
\midrule
Ours &\textbf{37.24} & \textbf{55.20} & \textbf{68.89} & \textbf{36.02} & \textbf{56.99} & \textbf{73.72} & \textbf{41.00} & \textbf{51.08}& \textbf{65.25} & \textbf{39.38} & \textbf{55.07}& \textbf{71.42}\\
\midrule
\bottomrule
\end{tabular}}
\end{table}
\begin{itemize}
    \item We generate synthetic raindrops featuring diverse sizes, intensities, and descent angles. This is accomplished by conceptualizing raindrops as ellipses characterized by variable degrees of transparency and blur to mimic motion. Consequently, we represent raindrops as an aggregation of ellipses.
    \begin{equation}
         R(x, y) = \left\{ (x_i, y_i, r_i^{l}, r_i^{s}, \theta_i, \alpha_i) \mid i = 1, \ldots, N \right\},
    \end{equation}
    where $(x_i, y_i)$ are the coordinates of the $i$-th raindrop, $r_i^{l}$ and $r_i^{s}$ denote the long radius and the short radius, $\theta_i$ is the falling angle, and $\alpha_i$ is the transparency.
    
    \item Then the generated raindrops are inserted onto each frame of the video. This involves blending the raindrop layer with the original video frames using beta blending, where the final pixel value $I'$ is given by:
    \begin{equation}
        I' = (1 - \beta_i)I + \beta_iR,
    \end{equation}
    $I$ is the original pixel intensity, and $R$ is the raindrop intensity.
    
    \item Finlay we apply random motion blur to the raindrops to simulate the falling motion. The extent of the blur corresponds to the speed and angle of the rain, enhancing the realism of the effect.
\end{itemize}
The parameters for raindrop generation, such as size, intensity, angle, and speed, are chosen and varied randomly to simulate natural rain. 
Additionally, the ambient lighting and camera effects, like reflections and refractions, can also be considered for a more realistic simulation.
The transparency $\alpha_i$ is chosen randomly from $\left[0.9, 0.8, 0.7, 0.6, 0.5\right]$. The long radius of the ellipse is chosen as $20$ pixels while the short radius of the ellipse is chosen as $1$ pixel. The position of the raindrop is randomly chosen among all the positions of one frame. The angle $\theta_i$ is chosen randomly in $\left[-10^{\circ},10^{\circ} \right]$. $\beta_i$ is randomly chosen from $\left[ 0,0.5\right]$. We choose $N=83$ for each frame in the perturbed videos.

\subsection{Fog Noise}
The fog noise is intended to replicate the visual phenomenon of fog, which is distinguished by diminished contrast, decreased saturation, and a progressive white overlay that intensifies with distance. The procedure for simulating fog within video frames can be executed as follows.
Given:
\begin{itemize}
    \item Map size $n = Image Size$,
    \item Wibble decay factor $d = 3$,
    \item Initial step size $s = n$,
    \item Initial wibble value $w = 100$.
\end{itemize}

The plasma fractal heightmap $H$ is initialized with dimensions $n$ and starting values. At each iteration:

\begin{itemize}
    \item \textbf{Square step}: For each square in the grid:
    \begin{equation}
        H_{i+\frac{s}{2}, j+\frac{s}{2}} = \frac{H_{i,j} + H_{i+s,j} + H_{i,j+s} + H_{i+s,j+s}}{4} + \Delta w, 
    \end{equation}
    where $\Delta w$ is a random value from $[-w, w]$ and $s$ is the current step size.
    
    \item \textbf{Diamond step}: For each diamond in the grid, we calculate the center value as the mean of the four corner points plus a random value:
    \begin{equation}
        H_{i, j} = \frac{H_{i-\frac{s}{2},j} + H_{i+\frac{s}{2},j} + H_{i,j-\frac{s}{2}} + H_{i,j+\frac{s}{2}}}{4} + \Delta w.
    \end{equation}
    \item Update the step size and wibble value:
    \begin{equation}
       s := \frac{s}{2}, \quad w := \frac{w}{d}.
    \end{equation}
\end{itemize}
This process repeats until the step size $s \geq 2$.

\subsubsection{Fog Effect Application.}

For a given image sequence, the fog effect is applied based on a severity level which determines constants $(C_1, C_2)$ from a predefined set. The fog layer $F$ is generated by multiplying the plasma fractal with constant $C_1$:
\begin{equation}
    F = C_1 \times PlasmaFractal(wibbledecay=C_2).
\end{equation}

Then, for each image $I$ in the sequence, the fog is applied as follows:

\begin{itemize}
    \item Scale the image $I$ to the range $[0, 1]$.
    \item Trim $I$ to the central region of interest.
    \item Add the fog layer $F$ to $I$, ensuring the fog does not exceed the original brightness.
    \item Apply normalization to maintain image contrast:
    \begin{equation}
         I_{\text{fog}} = \text{clip}\left(\frac{I \times \text{max}(I)}{\text{max}(I) + C_1}, 0, 1\right) \times 255.
    \end{equation}
\end{itemize}

The result is the fog-enhanced image sequence. We choose $C_1=1.5$ and $C_2=2.5$ to simulate the fog effect.

\subsection{Analysis of the Rain and Fog Noises} 
The experimental results by injecting rain and fog noises on the val and test sets are delivered in Tab.~\ref{tab:robust_1}. We conduct experiments on the four most outperforming baselines selected from our benchmark, \ie, Singularity~\cite{lei2021less}, XCLIP~\cite{ma2022x}, AskAnything~\cite{2023videochat}, BLIPv2~\cite{li2023BLIP}, and on our proposed method \texttt{RefAtomNet}.
We first observe that by injecting two types of noise, all the methods show performance decay, while the fog noise demonstrates more negative influence on the RAVAR performances compared with the rain noise, as fog noise will blur more detailed visual cues. 
\texttt{RefAtomNet} delivers the best performances by $37.24\%$, $55.20\%$, $68.89\%$ and $36.02\%$, $56.99\%$, $73.72\%$ of mIOU, mAP, and AUROC, on val and test sets respectively under test-time rain noise, while delivering $41.00\%$, $51.08\%$, $65.25\%$ and $39.38\%$, $55.07\%$, $71.42\%$ of mIOU, mAP, and AUROC, on val and test sets respectively under test-time fog noise.

\begin{table}[t]
\centering
\caption{Experimental results of the most outperforming baselines and \texttt{RefAtomNet} when shot noise and Gaussian noise perturbations are added into the videos in the test phase.}
\label{tab:robust_2}
\scalebox{0.8}{\begin{tabular}{l|ccc|ccc|ccc|ccc}

\toprule
\midrule
\multirow{3}{*}{\textbf{Method}} & \multicolumn{6}{c|}{\textbf{Test-time shot noise perturbation}} & \multicolumn{6}{c}{\textbf{Test-time Gaussian noise perturbation}} \\
\cmidrule{2-13}
& mIOU    & mAP    & AUROC   & mIOU   & mAP   & AUROC   & mIOU    & mAP    & AUROC   & mIOU   & mAP   & AUROC   \\
\cmidrule{2-13}
& \multicolumn{3}{c|}{Val}& \multicolumn{3}{c|}{Test} & \multicolumn{3}{c|}{Val} & \multicolumn{3}{c}{Test}\\
\midrule
Singularity~\cite{lei2021less} & 8.77 & 39.20& 55.85 & 9.26& 40.34 &53.11  & 6.28& 38.00 & 54.37& 6.15&40.36 &53.53 \\
XClip~\cite{ma2022x} & 30.65 &40.62 & 53.79& 29.47& 37.11& 54.64& 34.15& 40.38 & 53.63 & 31.48& 37.41&54.88 \\
AskAnything~\cite{2023videochat} &21.52 &48.35 &62.96  &23.29 &49.03  &66.26  & 20.80 & 47.54& 60.88 & 22.41& 47.26& 64.54\\
BLIPv2~\cite{li2023BLIP} &32.12 &50.31 & 64.19 &32.39 & 52.24 & 68.12 & 31.75&48.63 &62.08  &32.22 &49.81 & 66.13\\
\midrule
Ours &\textbf{37.44} &\textbf{52.24} &\textbf{66.30} &\textbf{36.35} &\textbf{54.71} &\textbf{70.79}  &\textbf{36.48} &\textbf{50.16} &\textbf{63.61}  &\textbf{34.78} &\textbf{52.08} &\textbf{68.79} \\
\midrule
\bottomrule
\end{tabular}}
\end{table}

\subsection{Shot Noise} 
Shot noise, alternatively known as Poisson noise, is characterized by fluctuations that conform to a Poisson distribution. This type of noise is predominantly associated with the quantized nature of electronic charges or photons in optical systems. The injection of shot noise into video frames aims to simulate the intrinsic randomness encountered in real camera sensors, which is a consequence of the quantum properties of light.
To incorporate shot noise into a video frame, execute the subsequent steps for each pixel within such frame:
\begin{itemize}
    \item Denote the original pixel value as $I$, which represents the mean number of photons (or intensity) detected.
    \item Generate a new pixel value $I'$, which is a random value drawn from the Poisson distribution with mean $I$. The new value can be represented as:
    \begin{equation}
        I' \sim \text{Poisson}(s*I) = \frac{e^{-(s*I)} (s*I)^k}{k!},
    \end{equation}
\end{itemize}
where $k$ is the actual observed count, and $s$ is the severity chosen as $5$. 

\subsection{Gaussian Noise} Gaussian noise is widely used in video data processing because it closely mimics the natural noise presented in electronic devices and sensors due to thermal motion and other factors. Additionally, its mathematical properties and ease of implementation make it a standard choice for benchmarking and testing video processing algorithms. To inject Gaussian noise into video frames, we follow the following steps for each pixel in each frame:

\begin{itemize}
    \item We first determine the desired noise level, which is typically characterized by the standard deviation $\sigma$ of the Gaussian distribution. The mean $\mu$ of the distribution is often set to $0$ for noise injection purposes.
    \item For each pixel in the frame with the original intensity value $I$, we generate a random value $n$ from a Gaussian distribution with mean $\mu$ and standard deviation $\sigma$. This procedure can be represented as:
    \begin{equation}
        n \sim \mathcal{N}(\mu, \sigma^2).
    \end{equation}
    \item Finally, we add the noise value $n$ to the original pixel value $I$ to obtain the new noisy pixel value $I'$:
    \begin{equation}
         I' = I + n.
    \end{equation}
\end{itemize}

The resulting noisy pixel values $I'$ might exceed the valid range of pixel values (\textit{e.g.}, $0$ to $255$ for an $8$-bit image). Therefore, it is common to clip the values to remain within this valid range after the noise has been added.
We choose $\mu=0$ and $\sigma=0.2$ in our experiments.

\subsection{Analysis of the Shot and Gaussian Noises} 
\begin{figure*}[t!]
\centering
\includegraphics[width=1\linewidth]{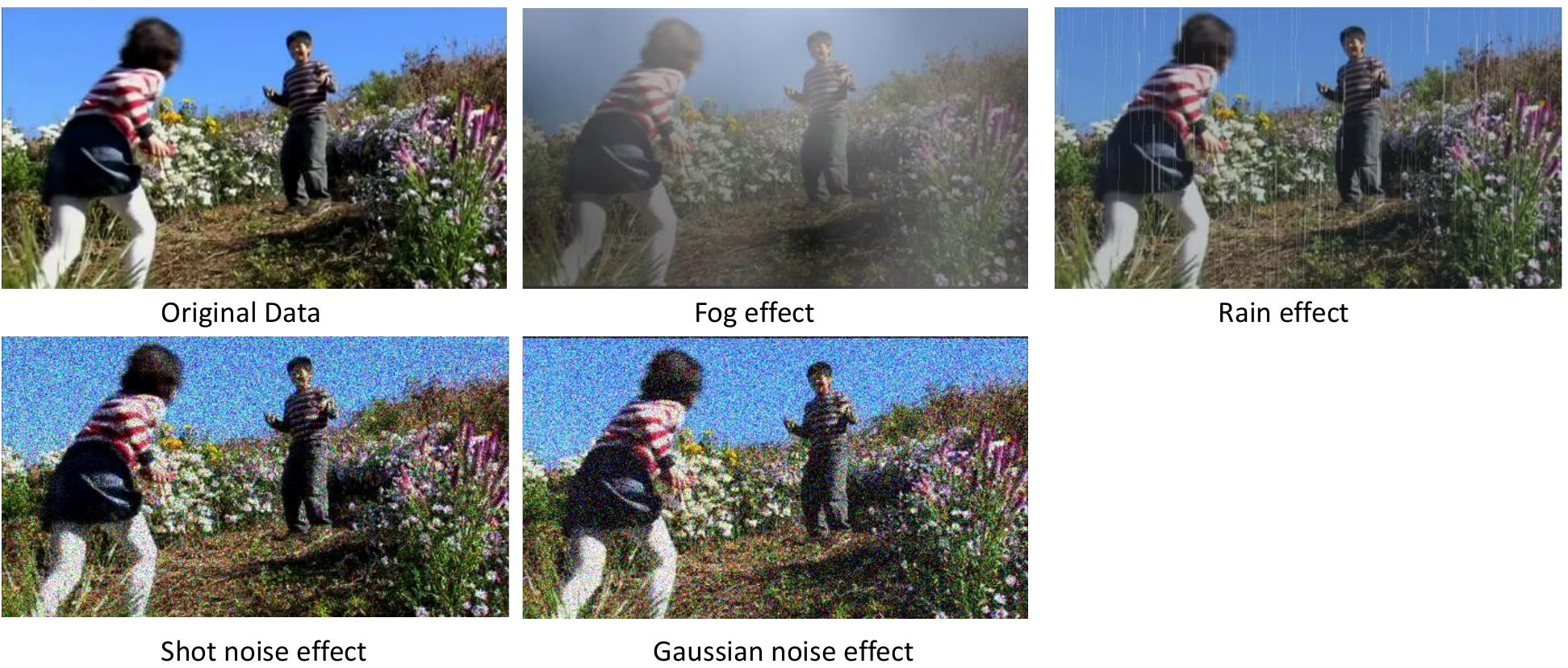}
\caption{An overview of a perturbed frame by using four different kinds of perturbation.}
\label{fig:perturb}
\end{figure*}

The experimental outcomes resulted from the application of shot and Gaussian noises on the val and test sets are revealed in Tab.~\ref{tab:robust_2}.
Experiments were conducted with the four highest-performing baselines identified from our benchmarks, namely, Singularity~\cite{lei2021less}, XCLIP~\cite{ma2022x}, AskAnything~\cite{2023videochat}, BLIPv2~\cite{li2023BLIP}, as well as our method \texttt{RefAtomNet}.
Generally, Gaussian noise was deemed to pose a greater challenge than shot noise due to its propensity to obscure finer visual details. Reflected in the results, the introduction of both noise types precipitates a decline in the performance across all evaluated methods, with Gaussian noise exerting a more detrimental impact on the RAVAR performances in comparison to shot noise.
In scenarios characterized by test-time shot noise, \texttt{RefAtomNet} won the competition, achieving mIOU, mAP, and AUROC scores of $37.44\%$, $52.24\%$, $66.30\%$ on the val set, and $36.35\%$, $54.71\%$, $70.79\%$ on the test set, respectively. Notably, under conditions of test-time Gaussian noise, \texttt{RefAtomNet} reported mIOU, mAP, and AUROC scores of $36.48\%$, $50.16\%$, $63.61\%$ on the val set, and $34.78\%$, $52.08\%$, $68.79\%$ on the test set, respectively. We further provide the visualizations of these four different kinds of perturbations in Fig.~\ref{fig:perturb}.

\begin{figure*}[t!]
\centering
\includegraphics[width=1\linewidth]{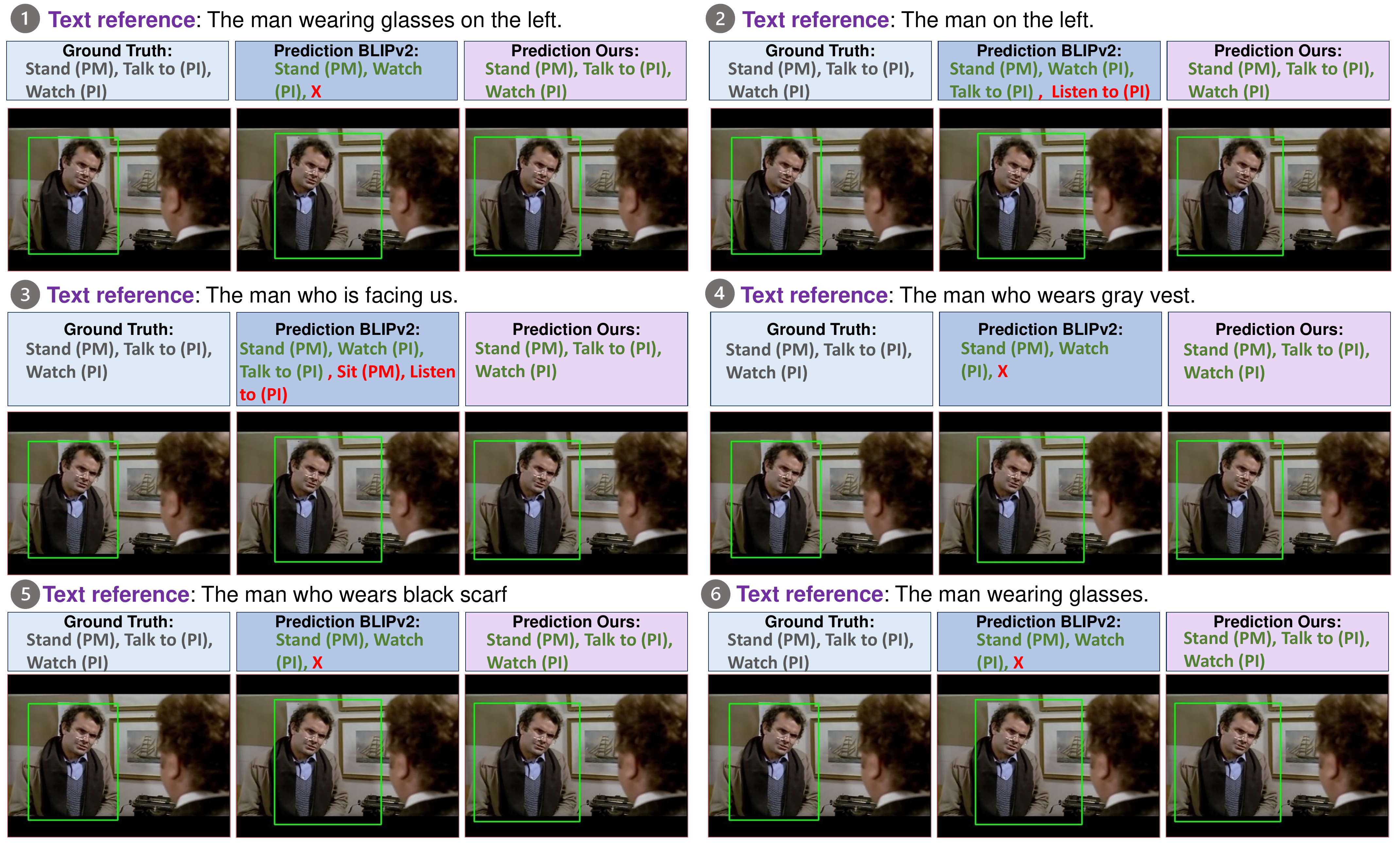}
\caption{Qualitative results for the test time rephrasing.}
\label{fig:rephrasing}
\end{figure*}
\begin{figure*}[t!]
\centering
\includegraphics[width=1\linewidth]{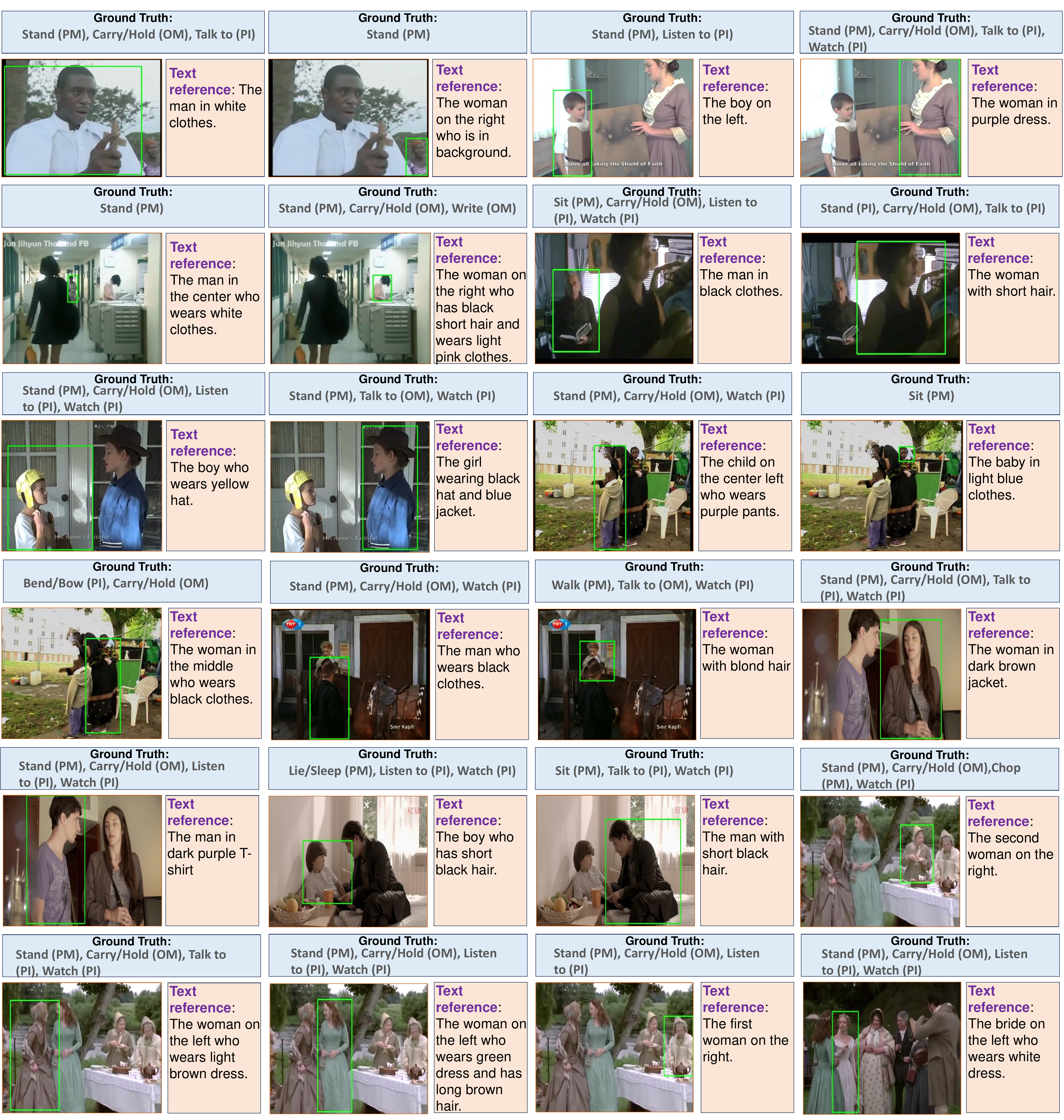}
\caption{More samples from our RefAVA dataset.}
\label{fig:samples}
\end{figure*}

\section{Qualitative Results for Test Time Rephrasing} 
We further deliver a sample towards test rephrasing, where we referred the person of interest with different descriptions in Fig.~\ref{fig:rephrasing}. We set the threshold as $0.91$ for both of the models to get the multi-label predictions. The person of interest is textually referred to as \textit{the man wearing glasses on the left}, \textit{the man on the left}, \textit{the man who is facing us}, \textit{the man who wears gray vest}, \textit{the man who wears black scarf}, and \textit{the man wearing glasses}, respectively. We observe that the predicted locations of the person do not change among different textual descriptions, varying from the visual appearance attributes leveraged for the indication. The atomic action recognition results of our \texttt{RefAtomNet} preserve consistency and deliver concrete predictions for the person of interest. However, there are small fluctuations in the atomic action predictions of the baseline BLIPv2~\cite{li2023BLIP}. 
These results demonstrate the strong generalizability of our approach towards the varied descriptions during the test time, which is essential for real-world applications since the textual descriptions may differ among different users according to the person's appearance attributes.

\section{More Samples of the RefAVA Dataset}
In this section, we deliver more samples from the contributed RefAVA dataset, as shown in Fig.~\ref{fig:samples}, where for each instance, the textual reference sentence is displayed in the light orange box on the right side of the image, the atomic action annotations and the bounding box are shown on the top of the image and within the image itself using a green box.

\end{document}